%% file: acl_latex.tex
\pdfoutput=1

\documentclass[11pt]{article}

\usepackage[]{acl}

\usepackage{times}
\usepackage{latexsym}

\usepackage[T1]{fontenc}

\usepackage[utf8]{inputenc}

\usepackage{microtype}

\usepackage[utf8]{inputenc} 
\usepackage[T1]{fontenc}    
\usepackage{hyperref}       
\usepackage{url}            
\usepackage{booktabs}       
\usepackage{amsfonts}       
\usepackage{nicefrac}       
\usepackage{microtype}      
\usepackage{xcolor}         
\usepackage{multirow}
\usepackage{CJKutf8}        
\usepackage[pdftex]{graphicx}
\nointerlineskip 
\usepackage{subfigure}
\usepackage{makecell}

\newcommand*\samethanks[1][\value{footnote}]{\footnotemark[#1]}

\input{header}

\definecolor{highlight}{rgb}{0,0,0} 
\definecolor{highlight}{rgb}{0,0,0} 

\title{CBLUE: A Chinese Biomedical Language Understanding Evaluation Benchmark}

\author{Ningyu Zhang$^{1}\thanks{~~Equal contribution and shared co-first authorship.}$, Mosha Chen$^{2}\samethanks$, Zhen Bi$^{1}\samethanks$, Xiaozhuan Liang$^{1}\samethanks$, Lei Li$^{1}\samethanks$, Xin Shang$^{3}$\\
\textbf{Kangping Yin$^{2}$, Chuanqi Tan$^{2}$, Jian Xu$^{2}$, Fei Huang$^{2}$, Luo Si$^{2}$, Yuan Ni$^{4}$, Guotong Xie$^{4,5,6}$ }\\
\textbf{Zhifang Sui$^{7,13}$, Baobao Chang$^{7,13}$, Hui Zong$^{8,14}$, Zheng Yuan$^{9}$, Linfeng Li$^{10}$, Jun Yan$^{10}$}\\
\textbf{Hongying Zan$^{11,13}$, Kunli Zhang$^{11,13}$, Buzhou Tang$^{12,13}\thanks{~~Corresponding author.}$, Qingcai Chen$^{12,13}\footnotemark[2]$}\\

$^{1}$Alibaba-Zhejiang University Joint Research Institute of Frontier Technologies, Zhejiang University \\
$^{2}$Alibaba Group, $^{3}$School of Mathematical Science, Zhejiang University, $^{4}$Pingan Health Technology \\
$^{5}$Ping An Health Cloud Company Limited 
$^{6}$Ping An International Smart City Technology Co., Ltd \\
$^{7}$Key Laboratory of Computational Linguistics, Ministry of Education, Peking University\\
$^{8}$School of Life Sciences and Technology, Tongji University 
$^{9}$Tsinghua University, \\
$^{10}$Yidu Cloud Technology Inc 
$^{11}$School of Information Engineering, Zhengzhou University \\
$^{12}$Harbin Institute of Technology (Shenzhen) 
$^{13}$Peng Cheng Laboratory, 
$^{14}$Philips Research China\\

}

\begin{document}
\maketitle
\begin{abstract}
With the development of biomedical language understanding benchmarks, Artificial Intelligence applications are widely used in the medical field. However, most benchmarks are limited to English, which makes it challenging to replicate many of the successes in English for other languages. To facilitate research in this direction, we collect real-world biomedical data and present the first Chinese Biomedical Language Understanding Evaluation (CBLUE) benchmark: a collection of natural language understanding tasks including named entity recognition, information extraction, clinical diagnosis normalization, 
and an associated online platform for model evaluation, comparison, and analysis. To establish evaluation on these tasks, we report empirical results with the current 11 pre-trained Chinese models, and experimental results show that state-of-the-art neural models perform far worse than the human ceiling\footnote{Code available in \url{https://github.com/CBLUEbenchmark/CBLUE}}. 
Our benchmark is released at \url{https://tianchi.aliyun.com/dataset/dataDetail?dataId=95414&lang=en-us}.

\end{abstract}

\section{Introduction}

Artificial intelligence is gradually changing the landscape of healthcare, and biomedical research \cite{yu2018artificial}. 
With the fast advancement of biomedical datasets, biomedical natural language processing (BioNLP) has facilitated a broad range of applications such as biomedical text mining, which leverages textual data in Electronic Health Records (EHRs).

A key driving force behind such improvements and rapid iterations of models is the use of general evaluation datasets and benchmarks \cite{gijsbers2019open}. 
Pioneer benchmarks, such as BLURB \cite{DBLP:journals/corr/abs-2007-15779}, PubMedQA \cite{DBLP:conf/emnlp/JinDLCL19}, and others, have provided us with the opportunity to conduct research on biomedical language understanding and developing real-world applications. 
Unfortunately, most of these benchmarks are developed in English, which makes the development of the associated machine intelligence Anglo-centric. 
Meanwhile, other languages, such as Chinese, have unique linguistic characteristics and categories that need to be considered.
Even though Chinese speakers account for a quarter of the world population, there have been no existing Chinese biomedical language understanding evaluation benchmarks. 

To address this issue and facilitate natural language processing studies in Chinese, we take the first step in introducing a comprehensive \textbf{C}hinese \textbf{B}iomedical \textbf{L}anguage \textbf{U}nderstanding \textbf{E}valuation (\textbf{CBLUE}) benchmark with eight biomedical language understanding tasks.
These tasks include named entity recognition, information extraction, clinical diagnosis normalization, short text classification, question answering (in transfer learning setting), intent classification, semantic similarity, and so on.
We evaluate several pre-trained Chinese language models on CBLUE and report their performance. 
The current models still perform by far worse than the standard of single-human performance, leaving room for future improvements.
We also conduct a comprehensive analysis using case studies to indicate the challenges and linguistic differences in Chinese biomedical language understanding. 
We intend to develop a universal GLUE-like open platform for the Chinese BioNLP community, and this work helps accelerate research in that direction.
Overall, the main contributions of this study are as follows:

\begin{table*}[!htbp]
  \centering
  \begin{tabular}{llllll}
    \toprule
    \textbf{Dataset} & \textbf{Task} & \textbf{Train} & \textbf{Dev} & \textbf{Test} & \textbf{Metrics} \\
    \midrule
    CMeEE & NER & 15,000 & 5,000 & 3,000 & Micro F1 \\
    CMeIE & Information Extraction & 14,339 & 3,585 & 4,482 & Micro F1 \\
    \midrule
    CHIP-CDN & Diagnosis Normalization & 6,000 & 2,000 & 10,192 & Micro F1 \\
    CHIP-STS & Sentence Similarity & 16,000 & 4,000 & 10,000 & Macro F1\\
    \midrule
    CHIP-CTC & Sentence Classification & 22,962 & 7,682 & 10,000 & Macro F1 \\
    KUAKE-QIC & Intent Classification & 6,931 & 1,955 & 1,994 & Accuracy \\
    \midrule 
    KUAKE-QTR & Query-Document Relevance & 24,174 & 2,913 & 5,465 & Accuracy\\
    KUAKE-QQR & Query-Query Relevance & 15,000 & 1,600 & 1,596 & Accuracy\\
    \bottomrule
  \end{tabular}
    \caption{Task descriptions and statistics in CBLUE. CMeEE and CMeIE are sequence labeling tasks. Others are single sentence or sentence pair classification tasks.
    }
    \label{tab:datasets}
\end{table*}

\begin{itemize}
    \item We propose the first Chinese biomedical language understanding benchmark, an open-ended, community-driven project with diverse tasks. 
    The proposed benchmark serves as a platform for the Chinese BioNLP community and encourages new dataset contributions.
    
    \item We report a systematic evaluation of 11 Chinese pre-trained language models to understand the challenges derived by these tasks. 
    We release the source code of the baselines as a toolkit for future research purposes.
    
\end{itemize}
 
\section{Related Work}

Several benchmarks have been developed to evaluate general language understanding over the past few years. 
GLUE \cite{DBLP:conf/iclr/WangSMHLB19} is one of the first frameworks developed as a formal challenge affording straightforward comparison between task-agnostic transfer learning techniques.
SuperGLUE \cite{DBLP:conf/nips/WangPNSMHLB19}, styled after GLUE, introduce a new set of more difficult language understanding datasets.
Other similarly motivated benchmarks include DecaNLP~\citep{DBLP:journals/corr/abs-1806-08730}, which recast a set of target tasks into a general question-answering format and prohibit task-specific parameters, and SentEval~\citep{DBLP:conf/lrec/ConneauK18}, which evaluate explicitly fixed-size sentence embeddings. 
Non-English benchmarks include RussianSuperGLUE \cite{DBLP:conf/emnlp/ShavrinaFESAMMT20} and CLUE \cite{DBLP:conf/coling/XuHZLCLXSYYTDLS20}, which is a community-driven benchmark with nine Chinese natural language understanding tasks. 
These benchmarks in the general domain provide a north star goal for researchers and are part of the reason we can confidently say we have made great strides in our field.

For BioNLP, many datasets and benchmarks have been proposed \cite{DBLP:journals/corr/abs-2004-10706,Li2016BioCreativeVC,Wu2019RENETAD} which promote the biomedical language understanding \cite{DBLP:conf/emnlp/BeltagyLC19,lewis-etal-2020-pretrained,lee2020biobert}. 
\citet{DBLP:journals/bmcbi/TsatsaronisBMPZ15} propose biomedical language understanding datasets as well as a competition on large-scale biomedical semantic indexing and question answering. 
\citet{DBLP:conf/emnlp/JinDLCL19} propose PubMedQA, a novel biomedical question answering dataset collected from PubMed abstracts.
\citet{DBLP:conf/lrec/PappasAP18} propose BioRead, which is a publicly available cloze-style biomedical machine reading comprehension (MRC) dataset. 
\citet{DBLP:journals/corr/abs-2007-15779} create a leaderboard featuring the Biomedical Language Understanding \& Reasoning Benchmark (BLURB). 
Unlike a general domain corpus, the annotation of a biomedical corpus needs expert intervention and is labor-intensive and time-consuming. 
Moreover, most of the benchmarks are based on English; ignoring other languages means that potentially valuable information may be lost, which can be helpful for generalization.

In this study, we focus on Chinese to fill the gap and aim to develop \textbf{the first Chinese biomedical language understanding benchmark}. 
Note that Chinese biomedical text is linguistically different from English and has its
domain characteristics, necessitating an evaluation BioNLP benchmark designed explicitly for Chinese.

\begin{table*}[!htbp]

  \centering
  \begin{tabular}{llllll}
    \toprule
    \textbf{Benchmark}  &  \textbf{Language} & \textbf{Domain} & \textbf{Data Distribution} & \textbf{Label Distribution}  \\
    \midrule
    CBLUE & Chinese  & medical& long-tailed (CMeEE) & non-i.i.d (CHIP-STS) \\
    CLUE & Chinese & general & uniform & i.i.d   \\
    BLURB & English & medical & uniform & i.i.d   \\
    \bottomrule
  \end{tabular}
    \caption{{\color{highlight} Difference between CBLUE, CLUE and BLURB. There are three major differences: a) CBLUE has a much more diverse task setting with different data sources in the biomedical domain including clinical trials, EHRs, medical forum, text books and search engine logs; b) CBLUE has a long-tailed distribution which is challenging; c) CBLUE contains a specific transfer learning scenario supported by the CHIP-STS dataset, in which the testing set has a different distribution fromthe training set.}}
      \label{tab:compare}
\end{table*}

\section{CBLUE Overview}

\subsection{Design Principle}

CBLUE consists of 8 biomedical language understanding tasks.
The task descriptions and statistics of CBLUE are shown Table \ref{tab:datasets}.
Unlike CLUE \cite{DBLP:conf/coling/XuHZLCLXSYYTDLS20} as shown in Table \ref{tab:compare}, CBLUE has a diverse data source (the annotation is expensive), richer task setting, thus, more challenging for NLP models.
We introduce the design principle of CBLUE as follows:

1) \emph{Diverse tasks}: CBLUE contain widespread token-level, sequence-level, sequence-pair tasks.

2) \emph{Variety of differently distributed data}: CBLUE collect data from various sources, including clinical trials, EHRs, medical forum, textbooks, and search engine logs with a real-world distribution.

3) \emph{Quality control in long-term maintenance}: We asked domain experts (doctors from Class A tertiary hospitals) to annotate datasets and carefully review data to ensure data quality.

\subsection{Tasks}

\paragraph{CMeEE} 
For this task, the dataset is first released in CHIP2020\footnote{\url{http://cips-chip.org.cn/}}.
Given a pre-defined schema, the task is to identify and extract entities from the given sentence and classify them into nine categories: disease, clinical manifestations, drugs, medical equipment, medical procedures, body, medical examinations, microorganisms, and department. 

\paragraph{CMeIE} 
For this task, the dataset is also released in CHIP2020 \cite{2020CMeIE}.
The goal of the task is to identify both entities and relations in a sentence following the schema constraints. 
There are 53 relations defined in the dataset, including 10 synonymous sub-relationships and 43 other sub-relationships.

\paragraph{CHIP-CDN} 
For this task, the dataset is to standardize the terms from the final diagnoses of Chinese electronic medical records.
Given the original phrase, the task is to normalize it to standard terminology based on the International Classification of Diseases (ICD-10) standard for Beijing Clinical Edition v601. 

\paragraph{CHIP-CTC} 
For this task, the dataset is to classify clinical trials eligibility criteria, which are fundamental guidelines of clinical trials defined to identify whether a subject meets a clinical trial or not \cite{DBLP:journals/midm/ZongYZLZ21}.
All text data are collected from the website of the Chinese Clinical Trial Registry (ChiCTR) \footnote{\url{http://chictr.org.cn/}}, and a total of 44 categories are defined. 
The task is like text classification; although it is not a new task, studies and corpora for the Chinese clinical trial criterion are \emph{still limited}, and we hope to promote future research for social benefits. 

\paragraph{CHIP-STS} 
For this task, the dataset is for sentence similarity in the non-i.i.d. (non-independent and identically distributed) setting.
Specifically, the task aims to evaluate the generalization ability between disease types on Chinese disease questions and answer data. 
Given question pairs related to 5 different diseases (The disease types in the training and testing set are different), the task is to determine whether the semantics of the two sentences are similar.

\paragraph{KUAKE-QIC} 
For this task, the dataset is for intent classification.
Given search engine queries, the task is to classify each of them into one of 11 medical intent categories defined in KUAKE-QIC.
Those include diagnosis, etiology analysis, treatment plan, medical advice, test result analysis and others.

\paragraph{KUAKE-QTR} 
For this task, the dataset is used to estimate the relevance of the title of a query document.
Given a query (e.g., ``Symptoms of vitamin B deficiency''), the task aims to find the relevant title (e.g., ``The main manifestations of vitamin B deficiency'').

\paragraph{KUAKE-QQR} 
For this task, the dataset is used to evaluate the relevance of the content expressed in two queries.
Similar to KUAKE-QTR, the task aims to estimate query-query relevance, which is an essential and challenging task in real-world search engines. 

\subsection{Data Collection} 
Since machine learning models are mostly data-driven, data plays a critical role, and it is pretty often in the form of a static dataset \cite{DBLP:journals/corr/abs-1803-09010}.
We collect data for different tasks from diverse sources, including clinical trials, EHRs, medical books, and search logs from real-world search engines. 
As biomedical data may contain private information such as the patient’s name, age, and gender, \textbf{all collected datasets are anonymized and reviewed by the IRB committee of each data provider to preserve privacy.}
We introduce the data collection details followingly.


\subsection*{Collection from Clinical Trials} 

Clinical trial eligibility criteria text is collected from ChiCTR, a non-profit organization that provides information about clinical trial registration for public research use.
In each trial registry file, eligibility criteria text is organized as a paragraph in the inclusion criteria and exclusion criteria.
Some meaningless texts are excluded, and the remaining texts are annotated to generate the CHIP-CTC dataset.

\subsection*{Collection from EHRs}

We obtain the final diagnoses of the medical records from several Class A tertiary hospitals and sample a few diagnosis items from different medical departments to construct the CHIP-CDN dataset for research purposes. 
{\color{highlight} The diagnosis items are randomly sampled from the items which are not covered by the common medical synonyms dict}. 
\textbf{No privacy information is involved in the final diagnoses.} 



\subsection*{Collection from Medical Forum and Textbooks} 

Due to the COVID-19 pandemic, online consultation has become more and more popular via the Internet. 
To promote data diversity, we select the online questions by patients to build the CHIP-STS dataset. 
Note that most of the questions are chief complaints. To ensure the authority and practicability of the corpus, we also select medical textbooks of Pediatrics \cite{2018Pediatrics}, Clinical Pediatrics \cite{2013ClinicalPediatrics} and Clinical Practice\footnote{\url{http://www.nhc.gov.cn/}}. 
We collect data from these sources to construct the CMeIE and CMeEE datasets.

\subsection*{Collection from Search Engine Logs}

We also collect search logs from real-world search engines like the Alibaba KUAKE Search Engine\footnote{\url{https://www.myquark.cn/}}.
First, we filter the search queries in the raw search logs by the medical tag to obtain candidate medical texts. 
Then, we sample the documents for each query with non-zero relevance scores (i.e., to determine if the document is relevant to the query). 
Specifically, we divide all the documents into three categories, namely high, middle, and tail documents,  and then uniformly sample the data to guarantee diversity. 
We leverage the data from search logs to construct  KUAKE-QTC, KUAKE-QTR, and KUAKE-QQR datasets.

\subsection{Annotation}

\begin{figure*}[!htbp]
\centering
\subfigure[CMeEE] { 
  \includegraphics[height=5.9cm]{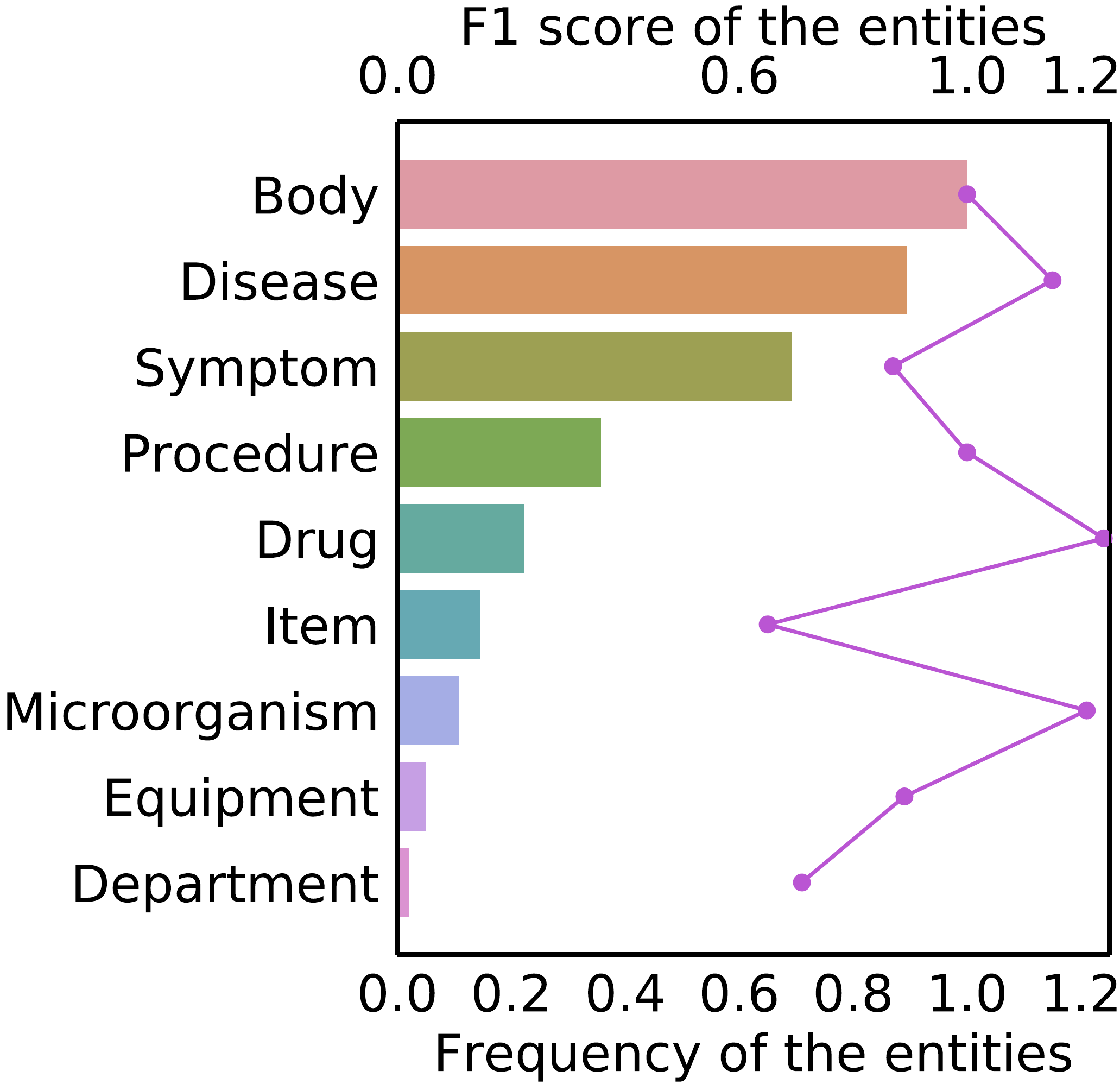}\label{f1}
}
\subfigure[CMeIE] {
\includegraphics[height=5.9cm]{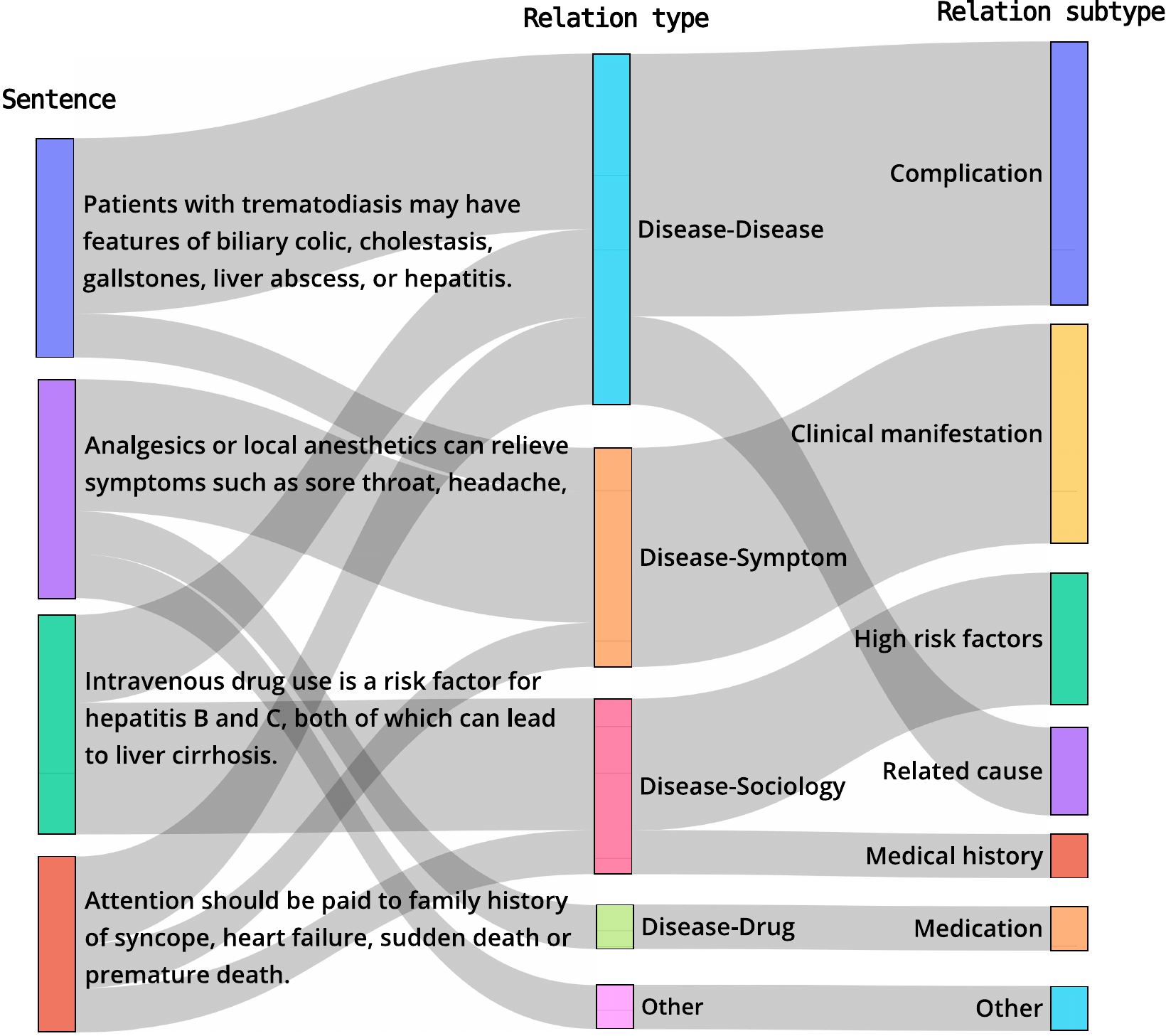}\label{f2}
}
\caption{
Analysis of the named entity recognition and information extraction datasets. 
(a) illustrates the entity (coarse-grained) distribution in CMeEE and the impact of data distribution on the model's performance. We set entity type Body with the maximum number of entities to 1.0, and others to the ratio of number or F1 score to Body. (b) shows the relation hierarchy in CMeIE.}

\label{fig:diverse}
\end{figure*}

Each sample is annotated by \textbf{three to five domain experts}, and the annotation with the majority of votes is taken to estimate human performance. 
During the annotation phase, we add control questions to prevent dishonest behaviors by the domain experts. 
Consequently, we reject any annotations made by domain experts who fail in the training phase and do not adopt the results of those who achieved low performance on the control tasks.
We maintain strict and high criteria for approval and review at least 10 random samples from each worker to decide whether to approve or reject all their HITs.
We also calculate the average inter-rater agreement between annotators using Fleiss' Kappa scores \cite{fleiss1971measuring}, finding that five out of six annotations show almost perfect agreement ($\kappa = 0.9$).

\subsection{Characteristics}

\paragraph{Utility-preserving Anonymization} 
Biomedical data may be considered as a breach in the privacy of individuals because they usually contain sensitive information.
Thus, we conduct utility-preserving anonymization following \cite{DBLP:journals/midm/LeeKKC17} to anonymize the data before releasing the benchmark. 

\paragraph{Real-world Distribution} 
To promote the generalization of models, all the data in our CBLUE benchmark follow real-world distribution without up/downsampling.
As shown in Figure \ref{f1}, our dataset follows long-tail distribution following Zipf’s law and will inevitably be long-tailed. However, long-tail distribution has no significant effect on performance.
Further, some datasets, such as CMedIE, have label hierarchy with both coarse-grained and fine-grained relation labels, as shown in Figure \ref{f2}.

\paragraph{Diverse Tasks Setting} 
Our CBLUE benchmark includes eight diverse tasks, including named entity recognition, relation extraction, and single-sentence/sentence-pair classification. 
Besides the independent and i.i.d. scenarios, our CBLUE benchmark also contains a specific \textbf{transfer learning} scenario supported by the CHIP-STS dataset, in which the testing set has a different distribution from the training set.

\subsection{Leaderboard}
We provide a leaderboard for users to submit their own results on CBLUE. 
The evaluation system will give final scores for each task when users submit their prediction results. 
The platform offers 60 free GPU hours from Aliyun\footnote{\url{https://tianchi.aliyun.com/notebook-ai/}} to help researchers develop and train their models.

\subsection{Distribution and Maintenance}
Our CBLUE benchmark was released online on April 1, 2021. 
Up to now, more than \textbf{300} researchers have applied the dataset, and over \textbf{80} teams have submitted their model predictions to our platform, including medical institutions (Peking Union Medical College Hospital, etc.), universities (Tsinghua University, Zhejiang University, etc.), and companies (Baidu, JD, etc.). 
We will continue to maintain the benchmark by attending to new requests and adding new tasks. 

\begin{table*}[!htbp]
\small
\centering
\begin{tabular}{l|cccccccc|c}
    \toprule

\textbf{Model} & \textbf{CMeEE} & \textbf{CMeIE} & \textbf{CDN} & \textbf{CTC} &  \textbf{STS} &  \textbf{QIC}  & \textbf{QTR}  & \textbf{QQR} & \textbf{Avg.}\\
    \midrule
 
    BERT-base & 62.1 & 54.0 & 55.4 & 69.2 & 83.0 & 84.3 & 60.0 & \textbf{84.7} &  69.1\\
    BERT-wwm-ext-base & 61.7 & 54.0 & 55.4 & 70.1 & 83.9 & 84.5 & 60.9 & 84.4 & 69.4\\
    RoBERTa-large  & 62.1 & 54.4 & 56.5 & \textbf{70.9} & 84.7 & 84.2 & 60.9 & 82.9 & 69.6\\
    RoBERTa-wwm-ext-base   & 62.4 & 53.7 & 56.4 & 69.4 & 83.7 & 85.5 & 60.3 & 82.7 & 69.3 \\
    RoBERTa-wwm-ext-large  & 61.8 & \textbf{55.9} & 55.7 & 69.0 & 85.2 & 85.3 & 62.8 & 84.4 & 70.0 \\
    ALBERT-tiny    & 50.5 & 35.9 & 50.2 & 61.0 & 79.7 & 75.8 & 55.5 & 79.8 & 61.1\\
    ALBERT-xxlarge  & 61.8 & 47.6 & 37.5 & 66.9 & 84.8 & \textbf{84.8} & 62.2 & 83.1 & 66.1\\
    ZEN & 61.0 & 50.1 & 57.8 & 68.6 & 83.5 & 83.2 & 60.3 & 83.0 & 68.4 \\
    MacBERT-base & 60.7 & 53.2 & 57.7 & 67.7 & 84.4 & 84.9 & 59.7 & 84.0 & 69.0 \\
    MacBERT-large & \textbf{62.4} & 51.6 & \textbf{59.3} & 68.6 & \textbf{85.6} & 82.7 & \textbf{62.9} & 83.5 & 69.6 \\
    PCL-MedBERT   & 60.6 & 49.1 & 55.8 & 67.8 & 83.8 & 84.3 & 59.3 & 82.5 & 67.9\\

    \midrule
    Human  & 67.0 & 66.0 & 65.0 & 78.0 & 93.0 & 88.0 & 71.0 & 89.0 & 77.1\\
    \bottomrule
\end{tabular}
\caption{Performance of baseline models on CBLUE benchmark.
}
\label{benchmark}
\end{table*}

\subsection{Reproducibility}

To make it easier to use the CBLUE benchmark, we also offer a toolkit implemented in PyTorch \cite{paszke2019pytorch} for reproducibility. 
Our toolkit supports mainstream pre-trained models and a wide range of target tasks. 

\section{Experiments}

\paragraph{Baselines} 
We conduct experiments with baselines based on different Chinese pre-trained language models. 
We add an additional output layer (e.g., MLP) for each CBLUE task and fine-tune the pre-trained models. 

\paragraph{Models} 
We evaluate CBLUE on the following public available Chinese pre-trained models:
 
\begin{itemize}
    \item 
    BERT-base \cite{devlin2018bert}. We use the base model with 12 layers, 768 hidden layers, 12 heads, and 110 million parameters.
    \item 
    BERT-wwm-ext-base \cite{cui2019pre}. A Chinese pre-trained BERT model with whole word masking. 
    \item 
    RoBERTa-large \cite{liu2019roberta}. Compared with BERT, RoBERTa removes the next sentence prediction objective and dynamically changes the masking pattern applied to the training data.
    \item 
    RoBERTa-wwm-ext-base/large. RoBERTa-wwm-ext is an efficient pre-trained model which integrates the advantages of RoBERTa and BERT-wwm.
    \item
    ALBERT-tiny/xxlarge \cite{lan2019albert}. ALBERT is a pre-trained model with two objectives: Masked Language Modeling (MLM) and Sentence Ordering Prediction (SOP). 
    \item 
    ZEN \cite{diao2019zen}. A BERT-based Chinese text encoder enhanced by N-gram representations, where different combinations of characters are considered during training.
    \item 
    Mac-BERT-base/large \cite{cui2020revisiting}. Mac-BERT is an improved BERT with novel MLM as a correction pre-training task.
    \item 
    PCL-MedBERT\footnote{\url{https://code.ihub.org.cn/projects/1775}}. A pre-trained medical language model proposed by the Peng Cheng Laboratory. 
    
\end{itemize}

\begin{table*}[!htbp]
\small
\centering
\begin{tabular}{l | lllllllll}

\toprule

& & \textbf{CMeEE} & \textbf{CMeIE} & \textbf{CDN} & \textbf{CTC} & \textbf{STS}  & \textbf{QIC} & \textbf{QTR} & \textbf{QQR}\\

\midrule

\multirow{5}{*}
{
    \begin{tabular}[c]{@{}l@{}}
        \textbf{Trained} \\
        \textbf{annotation}
    \end{tabular}
} 
& annotator 1  & 69.0 & 62.0 & 60.0 & 73.0 & 94.0 & 87.0 & 75.0 & 80.0 \\
& annotator 2  & 62.0 & 65.0 & 69.0 & 75.0 & 93.0 & 91.0 & 62.0 & 88.0 \\ 
& annotator 3  & 69.0 & 67.0 & 62.0 & 80.0 & 88.0 & 83.0 & 71.0 & 90.0  \\ 

\cmidrule{2-10}

& avg          & 66.7 & 64.7 & 63.7 & 76.0 & 91.7 & 87.0 & 69.3 & 86.0 \\ 
& majority & \textbf{67.0} &\textbf{66.0} & \textbf{65.0} & \textbf{78.0} & \textbf{93.0} & \textbf{88.0} & \textbf{71.0} & \textbf{89.0}   \\ 

\midrule


& best model & 62.4 & 55.9 & 59.3 & 70.9 & 85.6 & 85.5 & 62.9 & 84.7  \\ 
    \bottomrule
\end{tabular}
\caption{
Human performance of two-stage evaluation scores with the best-performed model. 
“avg” refers to the mean score from the three annotators.
“majority” indicates the performance taken from the majority vote {\color{highlight} of amateur humans}.
Bold text denotes the best result among human and model prediction.}
\label{human}
\end{table*}

\begin{figure*}[t]
\centering
\subfigure[Error Analysis in CMeEE] { 
  \includegraphics[height=2.8cm]{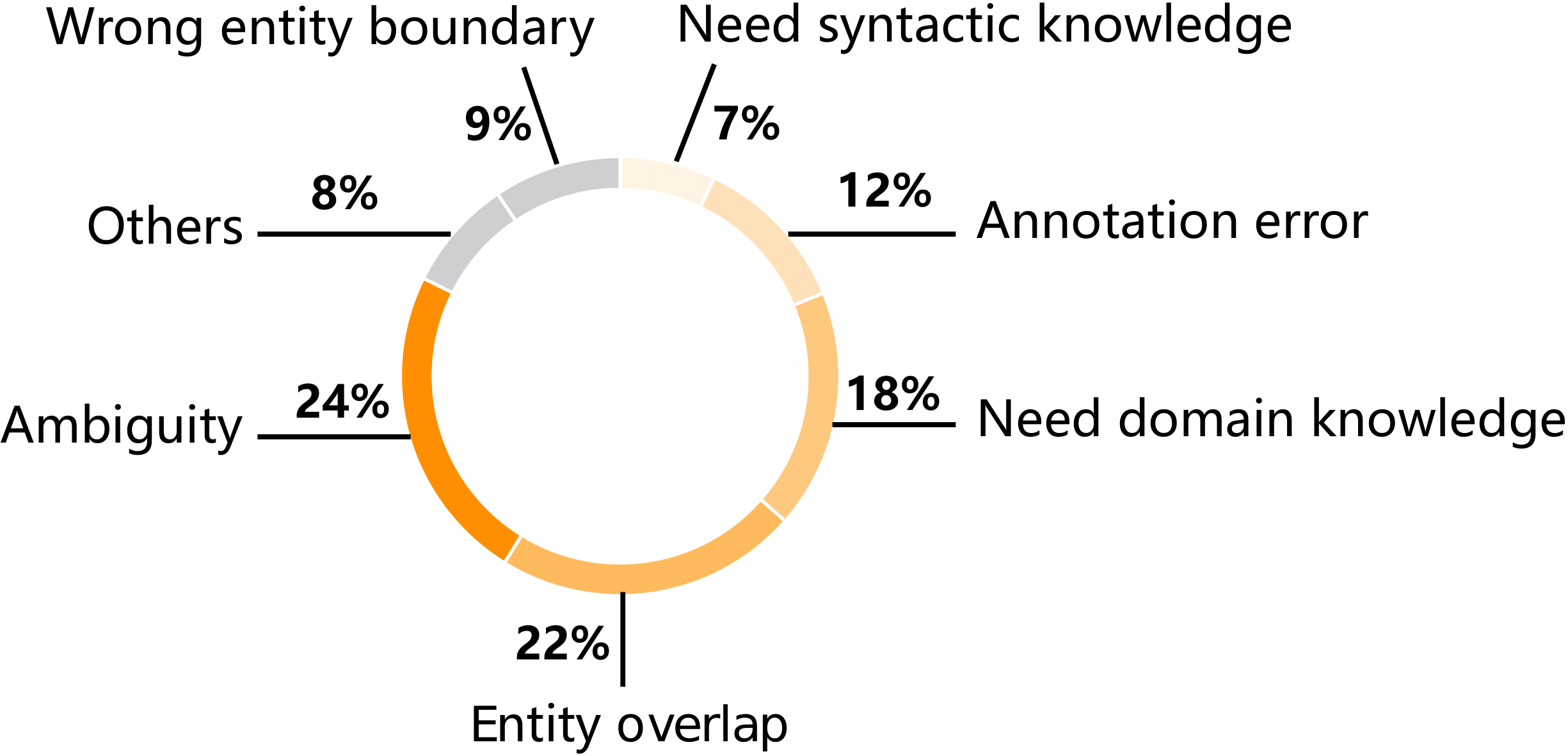}\label{e1}
}
\subfigure[Error Analysis in KUAKE-QIC] { 
    \includegraphics[height=2.9cm]{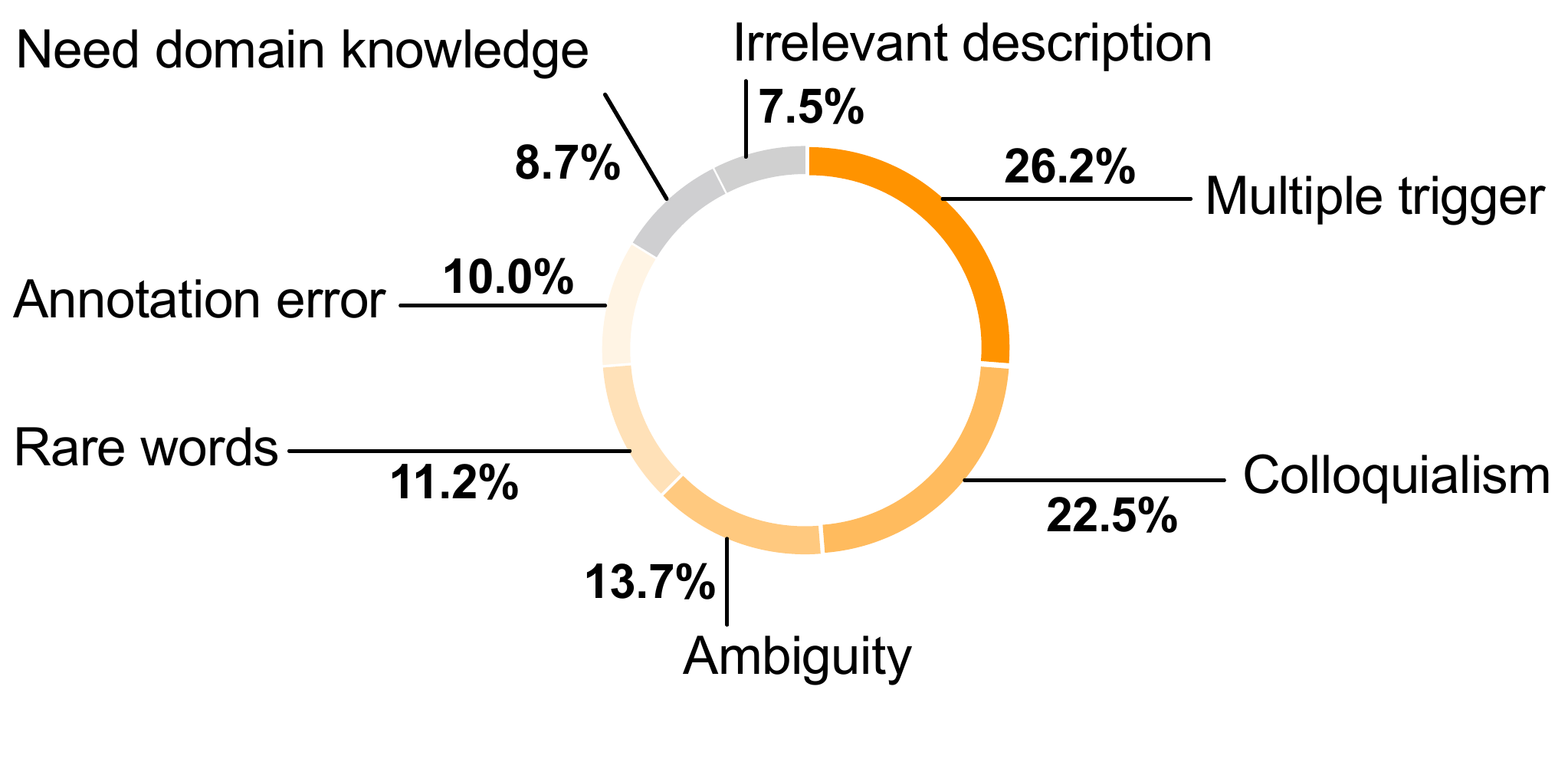}\label{e2}
}
\caption{
We conduct error analysis on datasets CMeEE and QIC. 
For CMeEE, we divide error cases into 6 categories, including ambiguity, need domain knowledge, entity overlap, wrong entity boundary, annotation error, and others (long sequence, rare words, etc.). 
For KUAKE-QIC, we divide error cases into 7 categories, including multiple triggers, colloquialism, ambiguity, rare words, annotation error, irrelevant description, and need domain knowledge.}
\label{exp2}
\end{figure*}

We implement all baselines with PyTorch \cite{paszke2019pytorch}.
All the training details can be found in the appendix.

\subsection{Benchmark Results}

{\color{highlight}
We report the results of our baseline models on the CBLUE benchmark in Table \ref{benchmark}. 
We notice that larger pre-trained models obtain better performance.
Since Chinese text is composed of terminologies, carefully designed masking strategies may be helpful for representation learning.
However, we observe that models which use whole word masking do not always yield better performance than others in some tasks, such as CTC, QIC, QTR, and QQR, indicating that tasks in our benchmark are challenging and more sophisticated technologies should be developed. 
Further, we find that ALBERT-tiny achieves comparable performance to base models in CDN, STS, QTR, and QQR tasks, illustrating that smaller models may also perform well in specific tasks.
We think this is caused by the different distribution between pretraining corpus and Chinese medical text; thus, large PTLMs may not obtain satisfactory performance. 
Finally, we notice that PCL-MedBERT, which tends to be state-of-the-art in Chinese biomedical text processing tasks, and does not perform as well as we expected.
This further demonstrates the difficulty of our benchmark, and contemporary models may find it difficult to quickly achieve outstanding performance.
}

\subsection{Human Performance}

For all of the tasks in CBLUE, we ask \textbf{human amateur annotators with no medical experience} to label instances from the testing set and compute the annotators’  majority vote against the gold label {\color{highlight} annotated by specialists}. 
Similar to SuperGLUE \cite{DBLP:conf/nips/WangPNSMHLB19}, we first need to train the annotators before they work on the testing data. 
Annotators are asked to annotate some data from the development set; then, their annotations are validated against the gold standard.
Annotators need to correct their annotation mistakes repeatedly so that they can master the specific tasks. 
Finally, they annotate instances from the testing data, and these annotations are used to compute the final human scores. 
The results are shown in Table \ref{human} and the last row of Table \ref{benchmark}. 
In all tasks, humans have better performance.

\subsection{Case studies} 
\begin{CJK}{UTF8}{gbsn}
{\color{highlight}
We choose two datasets: CMeEE and KUAKE-QIC, a sequence labeling and classification task, respectively, to conduct case studies. 
As shown in Figure \ref{exp2}, we report the statistics of the proportion of various types of error cases\footnote{See definitions of errors in the appendix.}. 
For CMeEE, we notice that \textit{entity overlap}\footnote{There exist multiple overlapping entities in the instance.}, \textit{ambiguity}\footnote{The instance has a similar context but different meaning, which mislead the prediction.}, \textit{need domain knowledge}\footnote{There exist biomedical terminologies in the instance which require domain knowledge to understand. }, \textit{annotation error}\footnote{The annotated label is wrong. } are major reasons that result in the prediction failure. 
Furthermore, there exist many instances with \textit{entity overlap}, which may lead to confusion for the named entity recognition task.
While in the analysis for KUAKE-QIC, almost half of bad cases are due to \textit{multiple triggers}\footnote{There exist multiple indicative words which mislead the prediction.} and \textit{colloquialism}. 
\textit{Colloquialism}\footnote{The instance is quite different from written language (e.g., with many abbreviations)} is natural in search queries, which means that some descriptions of the Chinese medical text are too simplified, colloquial, or inaccurate.

\begin{CJK*}{UTF8}{gbsn}
\begin{table*}[!htb]
  \footnotesize
  \label{tab:ee_badcase}
  \centering
  \setlength\tabcolsep{4pt}
  \begin{tabular}{p{5.7cm}p{3.2cm}p{1.6cm}<{\centering}p{0.3cm}<{\centering}p{1.6cm}<{\centering}}
    \toprule
    {\textbf{Sentence}} &
    {\textbf{Word}} &
    {\textbf{Label}} & 
    {\textbf{RO}} & 
    {\textbf{MB}} \\
    \midrule
    血液生化分析的结果显示维生素B缺乏率约为12\%～19\%。& 血液生化分析 & Ite & Pro & Pro\\
    The results of blood biochemical analysis show that vitamin B lack rate is about 12\% to 19\%. & blood biochemical analysis & Ite & Pro & Pro\\
    \midrule
    皮疹可因宿主产生特异性的抗毒素抗体而减少。 & 抗毒素抗体 & Bod & O & Bod\\
    The rash can be reduced by the host producing specific anti-toxin antibodies. & anti-toxin antibodies & Bod & O & Bod \\
    \midrule
    根据遗传物质的结构和功能改变的不同，可将遗传病分为五类：1.染色体病指染色体数目异常，或者染色体结构异常，包括缺失、易位、倒位等 & 缺失, 易位, 倒位 & Sym, Sym, Sym & O & Sym, Sym, Sym\\
    According to the structure and function of genetic material, genetic diseases are divided into five categories: 1. Chromosomal diseases refer to abnormal chromosome number or chromosome structure abnormalities, including deletions, translocations, inversions... &  deletions, translocations, inversions & Sym, Sym, Sym & O & Sym, Sym, Sym\\
    \bottomrule
  \end{tabular}
  \caption{
  Case studies in CMeEE. 
  We evaluate roberta-wwm-ext and PCL-MedBERT on 3 sampled sentences, with their gold labels and model predictions. 
  Ite (medical examination items), Pro (medical procedure), Bod (body), and Sym (clinical symptoms) are labeled for medical named words.
  O means that the model fails to extract the entity from sentences. RO=roberta-wwm-ext, MB=PCL-MedBERT.}
  \label{cmedee}
\end{table*}
\end{CJK*}

\begin{CJK*}{UTF8}{gbsn}
\begin{table*}[!htb]
  \footnotesize
  \label{tab:qqr_badcase}
  \centering
  \setlength\tabcolsep{4pt}
  \begin{tabular}{p{6cm} |  ccp{1.5cm}<{\centering} | p{1.5cm}<{\centering}}
    \toprule
    \multirow{2}{*}{\textbf{Query}} &
    \multicolumn{3}{c|}{\textbf{Model}} &
    \multirow{2}{*}{\textbf{Gold}} \\
    \cmidrule{2-4}
    & BERT & BERT-ext & MedBERT & \\
    \midrule
    请问淋巴细胞比率偏高、中性细胞比率偏低有事吗？ & 病情诊断 & 病情诊断 & 指标解读 & 指标解读 \\
    Does it matter if the ratio of lymphocytes is high and the ratio of neutrophils is low? & Diagnosis & Diagnosis & Test results analysis & Test results analysis \\
    \midrule
    咨询：请问小孩一般什么时候出水痘？ & 其他 & 其他 & 其他 & 疾病表述 \\
    Consultation: When do children usually get chickenpox? & Other & Other & Other & Disease description \\
    \midrule
    老人收缩压160，舒张压只有40多，是什么原因？怎么治疗？ & 病情诊断 & 病情诊断 & 病情诊断 & 治疗方案 \\
    The systolic blood pressure of the elderly is 160, and the diastolic blood pressure is only more than 40. What is the reason? How to treat? & Diagnosis & Diagnosis & Diagnosis & Treatment \\
    \bottomrule
  \end{tabular}
  \caption{
  Case studies in  KUAKE-QIC. 
  We evaluate the performance of baselines with 3 sampled instances. 
  The correlation between Query and Title is divided into 3 levels (0-2), which means `\textit{poorly related or unrelated}', `\textit{related}' and `\textit{strongly related}'. 
  BERT = BERT-base, BERT-ext = BERT-wwm-ext-base, MedBERT = PCL-MedBERT.}
  \label{qqr}
\end{table*}
\end{CJK*}

We show some cases on CMeEE in Table \ref{cmedee}.
In the second row, we notice that given the instance of  ``皮疹可因宿主产生特异性的抗毒素抗体而减少 (\textit{Rash can be reduced by the host producing specificanti-toxin antibodies.})'', ROBERTA and PCL-MedBERT obtain different predictions.
The reason is that there exist medical terms such as ``抗毒素抗体 (\textit{anti-toxin antibodies})''. 
ROBERTA can not identify those tokens correctly, but PCL-MedBERT, pre-trained on the medical corpus, can successfully make it.
Moreover, PCL-MedBERT can extract entities ``缺失,易位,倒位  (\textit{eletions, translocations, inversions})'' from the long sentences, which is challenging for other models.

We further show some cases on KUAKE-QIC in Table \ref{qqr}.
In the first case, we notice that both BERT and BERT-ext fail to obtain the intent label of the query ``请问淋巴细胞比率偏高、中性细胞比率偏低有事吗?  (\textit{Does it matter if the ratio of lymphocytes is high and the ratio of neutrophils is low?})'', while MedBERT can obtain the correct prediction. 
Since  ``淋巴细胞比率 (\textit{ratio of lymphocytes})'' and ``中性细胞比率 (\textit{ratio of neutrophils})'' are biomedical terms, and the general pre-trained language model has to leverage domain knowledge to understand those phrases. 
}

{
\color{highlight}
As shown in Table \ref{cmedee} and Table \ref{qqr}, compared with other languages, the Chinese language is very colloquial even in medical texts. 
Furthermore, polysemy is prevalent in chinese language.
The meaning of a word changes according to its tone, which usually causes confusion and difficulties for machine reading.  
}
In summary, we conclude that \textbf{tasks in CBLUE are not easy to solve since the Chinese language has unique characteristics}, and more robust models should be developed.


\end{CJK}

\section{Conclusion}

In this paper, we present a Chinese Biomedical Language Understanding Evaluation (CBLUE) benchmark. 
We evaluate 11 current language representation models on CBLUE and analyzed their results. 
The results illustrate the limited ability of state-of-the-art models to handle some of the more challenging tasks. 
In contrast to English benchmarks such as GLUE/SuperGLUE and BLURB, whose model performance already matches human performance, we observe that this is far from the truth for Chinese biomedical language understanding. 

\section*{Acknowledgments}
We want to express gratitude to the anonymous reviewers for their hard work and kind comments.
This work is funded by Special Project of New Generation Artificial Intelligence of the Ministry of Science and Technology of China (2021ZD0113402),
National Natural Science Foundations of China (61876052 and U1813215),
National Natural Science Foundation of Guangdong, China (2019A1515011158),
Strategic Emerging Industry Development Special Fund of Shenzhen (20200821174109001), Pilot Project in 5G + Health Application of Ministry of Industry and Information Technology \& National Health Commission (5G + Luohu Hospital Group: an Attempt to New Health Management Styles of Residents),
Zhengzhou collaborative innovation major special project (20XTZX11020),
Zhejiang Provincial Natural Science Foundation of China (No. LGG22F030011), 
Ningbo Natural Science Foundation (2021J190), 
and Yongjiang Talent Introduction Programme (2021A-156-G).

\section*{Ethical Considerations}

We collected all the data with authorization from the organization that owned the data and signed the agreement. 
We release the benchmark following the CC BY-NC 4.0 license.
All collected datasets are anonymized and reviewed by the  IRB  committee of each data provider to preserve privacy. 
Since we collect data following real-world distribution, there may exist popularity bias that cannot be ignored.



\bibliography{neurips_2021}
\bibliographystyle{acl_natbib}

\appendix

\begin{CJK*}{UTF8}{gbsn}

\section{Broader Impact}

The COVID-19 (coronavirus disease 2019) pandemic has had a significant impact on society, both because of the severe health effects of COVID-19 and the public health measures implemented to slow its spread. 
A lack of information fundamentally causes many difficulties experienced during the outbreak; attempts to address these needs caused an information overload for both researchers and the public. 
Biomedical natural language processing---the branch of artificial intelligence that interprets human language---can be applied to address many of the information needs making urgent by the COVID-19 pandemic. 
Unfortunately, most language benchmarks are in English, and no biomedical benchmark currently exists in Chinese.
Our benchmark CBLUE, as the first Chinese biomedical language understanding benchmark, can serve as an open testbed for model evaluations to promote the advancement of this technology.

{\color{highlight} \section{Negative Impact}

Although we ask domain experts and doctors to annotate all the corpus, there still exist some instances with wrong annotated labels. If a model was chosen based on numbers on the benchmark, this could cause real-world harm. Moreover,  our benchmark lowers the bar of entry to work with biomedical data. While generally a good thing, it may dilute the pool of data-driven work in the biomedical field even more than it already is, making it hard for experts to spot the relevant work.}

\section{Limitations}
\label{limit}
Although our CBLUE offers diverse settings, there are still some tasks not covered by the benchmark, such as medical dialogue generation \cite{DBLP:journals/corr/abs-2010-07497,DBLP:journals/corr/abs-2012-11988,DBLP:conf/emnlp/ZengYJYWZZZDZFZ20} or medical diagnosis \cite{DBLP:conf/acl/WeiLPTCHWD18}. 
We encourage researchers in both academics and industry to contribute new datasets. 
Besides, our benchmark is static; thus,  models may still achieve outstanding performance on tasks but fail on simple challenge examples and falter in real-world scenarios. 
We leave this as future works to construct a platform including dataset creation, model development, and assessment, leading to more robust and informative benchmarks.

\section{CBLUE Background}
\label{sec:appendix}

Standard datasets and shared tasks have played essential roles in promoting the development of AI technology.
Taking the Chinese BioNLP community as an example, the CHIP (China Health Information Processing) conference releases biomedical-related shared tasks every year, which has extensively advanced Chinese biomedical NLP technology.
However, some datasets are no longer available after the end of shared tasks, which has raised issues in the data acquisition and future research of the datasets.

In recent years, we can obtain state-of-the-art performance for many downstream tasks with the help of pre-trained language models.
A significant trend is the emergence of multi-task leaderboards, such as GLUE (General Language Understanding Evaluation) and CLUE (Chinese Language Understanding Evaluation).
These leaderboards provide a fair benchmark that attracts the attention of many researchers and further promotes the development of language model technology.
For example, Microsoft has released BLURB (Biomedical Language Understanding \& Reasoning Evaluation) at the end of 2020 in the medical field.
Recently, the Tianchi platform has launched the CBLUE (Chinese Biomedical Language Understanding Evaluation) public benchmark under the guidance of the CHIP Society.
We believe that the release of the CBLUE will further attract researchers' attention to the medical AI field and promote the development of the community.

CBLUE 1.0\footnote{We release the benchmark following the CC BY-NC 4.0 license.} comprises the previous shared tasks of the CHIP conference and the dataset from Alibaba QUAKE Search Engine, including named entity recognition, information extraction, clinical diagnosis normalization, single-sentence/sentence-pair classification.

\section{Detailed Task Introduction}

\subsection{Chinese Medical Named Entity Recognition Dataset (CMeEE)}

\paragraph{Task Background}
As an essential subtask of information extraction, entity recognition has achieved promising results in recent years. 
Biomedical texts such as textbooks, encyclopedias, clinical trials, medical literature, electronic health records, and medical examination reports contain rich medical knowledge. 
Named entity recognition is the process of extracting medical terminologies, such as diseases and symptoms, from the above mentioned unstructured or semi-structured texts, and it can help significantly improve the efficiency of scientific research.
CMeEE dataset is proposed for this purpose, and the original dataset was released at the CHIP2020 conference. 

\paragraph{Task Description}
This task is defined as given the pre-defined schema and an input sentence to identify medical entities and to classify them into 9 categories, including disease (dis), clinical symptoms(sym), drugs (dru), medical equipment (equ), medical procedures (pro), body (bod), medical examination items (ite), microorganisms (mic), department (dep).
{\color{highlight}  For the detailed annotation instructions, please refer to the CBLUE official website}, and examples are shown in Table \ref{appendix_cmeee}.

\begin{table*}[!htb]
    \begin{center}
    \centering
    
    \begin{tabular}{l l l p{5cm}}
    \toprule
    \textbf{Entity type} & \textbf{Entity subtype} & \textbf{Label} & \textbf{Example}  \\
    \toprule
    \makecell[l]{
        疾病\\
        disease
    } &
    \makecell[l]{
        疾病或综合症\\
        disease or syndrome\\
        中毒或受伤\\ 
        poisoned or injured\\
        器官或细胞受损\\
        damage to organs or cells
    } 
    & dis & 
    \makecell[l]{
        尿潴留者易继发泌尿系感染 \\
        Patients with urinary retention are\\
        prone to secondary infections of the\\
        urinary system. 
    }
    \\ 
    \midrule   
    \makecell[l]{
        临床表现\\   
        clinical manifestations
    }
    & 
    \makecell[l]{
        症状\\
        symptom\\
        体征\\
        physical sign
    }     
    & sym & 
    \makecell[l]
    {
        逐渐出现呼吸困难、阵发性喘憋，\\
        发作时呼吸快而浅，并伴有呼气\\
        性喘鸣，明显鼻扇及三凹征 \\
        Then dyspnea and paroxysmal\\
        asthma may occur, along with\\
        shortness of breath, expiratory\\
        stridor, obvious flaring nares,\\
        and three-concave sign.
    }
    \\
    \midrule    
    \makecell[l]{
        医疗程序\\
        medical procedure
    }
    &
    \makecell[l]{
        检查程序\\
        check procedure\\
        治疗 \\
        treatment \\
        或预防程序\\
        or preventive procedure
    }     
    & pro & 
    \makecell[l]{
        用免疫学方法检测黑种病原体的\\
        特异抗原很有诊断价值，因其简\\
        单快速，常常用于早期诊断，诊\\
        断意义常 较抗体检测更为可靠\\
        It is of great diagnostic value\\
        to detect the specific antigen of\\
        a certain pathogen with immunoassay,\\
        a simple and quick assay that is\\
        intended for early diagnosis and\\
        proves more reliable than the\\
        antibody assay.
    } 
    \\
    
    \bottomrule
    \end{tabular}
    
    
    \end{center}
    \caption{
    Examples in CMeEE
    }
    \label{appendix_cmeee}
\end{table*}

{\color{highlight} \paragraph{Annotation Process}

The annotation guide was conducted by two medical experts from Class A tertiary hospitals and optimized during the trail annotation process. A total of 32 annotators had participated in the annotation process, including 2 medical experts who are also the owner of the annotation guideline, 4 experts from the biomedical informatics field, 6 medical M.D., and 22 master students from computer science majors. The annotation lasts for about three months (from October 2018 to December 2018), as well as an additional month's time for curation. The total expense is about 50,000 RMB.

The annotation process was divided into two stages. 
\begin{itemize}
    \item Stage1: This stage was called the trail annotation phase. The medical experts gave training to the annotators to make sure they had a comprehensive understanding of the task. Two rounds of trail annotation were conducted by the annotators, with the purpose of getting familiar with the annotation task as well as discovering the unclear points of the guideline, and annotation problems were discussed, and the medical experts improved the annotation guidelines according to the feedback iteratively.
    \item Stage2: For the first phase, each record was assigned to two annotators to label independently, and the medical experts and biomedical informatics experts would give in time help. The annotation results were compared automatically by the annotation tools (developed for CMeEE and CMeIE tasks), and any disagreement was recorded and handed over to the next phase. In the second phase, medical experts and the annotators had a discussion for the disagreements records as well as other annotation problems, and the annotators made corrections. After the two stages, the IAA score (Kappa score) is 0.8537, which satisfied the research goal.
\end{itemize}

}

{\color{highlight} \paragraph{PII and IRB}
The corpus is collected from authorized medical textbooks or Clinical Practice, and no personally identifiable
information or offensive content is involved in the text. 

No PII is included in the above-mentioned resources. The dataset does not refer to ethics, which has been checked by the IRB committee of the provider.

The original dataset format is a self-defined plain text format. To simplify the data pre-processing step, the CBLUE team has converted the data format to the unified JSON format with the permission of the data provider.

}




\paragraph{Evaluation Metrics}
This task uses strict Micro-F1 metrics.

\paragraph{Dataset Statistic}
This task has 15,000 training set data, 5,000 validation set data, 3,000 test set data.
The corpus contains 938 files and 47,194 sentences. The average number of words contained per file is 2,355.
The dataset contains 504 common pediatric diseases, 7,085 body parts, 12,907 clinical symptoms, and 4,354 medical procedures in total.



\paragraph{Dataset Provider}
The dataset is provided by:
\begin{itemize}
    \item Key Laboratory of Computational Linguistics, Peking University, Ministry of Education, China
    \item Laboratory of Natural Language Processing, Zhengzhou University, China
    \item The Research Center for Artificial Intelligence, Peng Cheng Laboratory, China 
    \item Harbin Institute of Technology, Shenzhen, China
\end{itemize}

\subsection{Chinese Medical Information Extraction Dataset (CMeIE)}
\paragraph{Task Background}

Entity and relation extraction is an essential information extraction task for natural language processing and knowledge graph (KG), which is used to detect pairs of entities and their relations from unstructured text.
The technology of this task can apply to the medical field. 
For example,  with entity and relation extraction,  unstructured and semi-structured medical texts can construct medical knowledge graphs, which can serve lots of downstream tasks.

\paragraph{Task Description}
Given the schema and sentence, in which  defines the relation (Predicate) and its related Subject and Object, such as (``subject\_type'': ``疾病''，``predicate'': ``药物治疗''，``object\_type'': ``药物'').
The task requires the model to automatically analyze the sentence and then extract all the $ Triples = [(S1, P1, O1), (S2, P2, O2) ...] $ in the sentence.
Table \ref{appendix_cmeie} shows the examples in the data set, and 53 SCHEMAs include 10 kinds of genus relations, 43 other sub-relations.
The details are in the 53\_schema.json file.
{\color{highlight}  For the detailed annotation instructions, please refer to the CBLUE official website}, and examples are shown in Table \ref{appendix_cmeie}.


\begin{table*}[!htb]
    \begin{center}
    \centering
    
    \begin{tabular}{l l p{7.5cm}}
    \toprule
    \textbf{Relation type} & \textbf{Relation subtype}  & \textbf{Example}  \\
    \toprule
    \multirow{3}{*}
        {
        \makecell[l]{
        疾病\_其他\\
        disease\_other
        }
        } &
        \makecell[l]{
        预防\\
        prophylaxis
        }
        & 
        \{'predicate': '预防-prevention', 'subject': '麻风病-Leprosy', 'subject\_type': '疾病-disease', 'object': '利福-rifampicin', 'object\_type': '其他-others'\} 
        \\
        & 
        \makecell[l]{
        阶段\\
        phase
        }
        &
        \{'predicate': '阶段-phase', 'subject': '肿瘤-tumor', 'subject\_type': '疾病-disease', 'object': 'I期-phase\_$Ⅰ$', 'object\_type': '其他-others'\}
        \\ 
        &
        \makecell[l]{
        就诊科室\\
        treatment department
        }
        &
        \{'predicate': '就诊科室-treatment\_department', 'subject': '腹主动脉瘤-abdominal\_aortic\_aneurysm', 'subject\_type': '疾病-disease', 'object': '初级医疗保健医处-primary\_medical\_care\_clinic', 'object\_type': '其他-others'\}
      \\ 
    \midrule    
    \multirow{3}{*}
        {
        \makecell[l]{
        疾病\_其他治疗\\
        disease\_other treatment
        }    
        }
        &
        \makecell[l]{
        辅助治疗\\
        adjuvant therapy
        }
        &
        \{'predicate': '辅助治疗-adjuvant\_therapy', 'subject': '皮肤鳞状细胞癌-utaneous\_squamous\_cell\_carcinoma', 'subject\_type': '疾病-disease', 'object': '非手术破坏-non\_surgical\_destructio', 'object\_type': '其他治疗-other\_treatment'\} \\ 
        &
        \makecell[l]{
        化疗\\
        chemotherapy
        }
        & 
        \{'predicate': '化疗-chemotherapy', 'subject': '肿瘤-tumour', 'subject\_type': '皮肤鳞状细胞癌-cutaneous\_squamous\_cell\_carcinoma', 'object': '局部化疗-local\_chemotherapy', 'object\_type': '其他治疗-other\_treatment'\}
        \\ 
        &
        \makecell[l]{
        放射治疗\\
        radiotherapy
        }
        &
        \{'predicate': '放射治疗-radiation\_therapy', 'subject': '非肿瘤性疼痛-non\_cancer\_pain', 'subject\_type': '疾病-disease', 'object': '外照射-external\_irradiation', 'object\_type': '其他治疗-other\_treatment'\}
     \\
    \midrule    
    \makecell[l]{
        疾病\_手术治疗\\
        disease\_surgical treatment
    }
    &
    \makecell[l]{
        手术治疗\\
        surgical treatment
    }
     & \{'predicate': '手术治疗-surgical\_treatment', 'subject': '皮肤鳞状细胞癌-cutaneous 
     \_squamous\_cell\_carcinoma', 'subject\_type': '疾病-disease', 'object': '传统手术切除-surgical\_resection(traditional\_therapy)', 'object\_type': '手术治疗-surgical\_treatment'\} \\
    
    \bottomrule
    \end{tabular}
    
    
    \end{center}
    \caption{
    Examples in CMeIE
    }
    \label{appendix_cmeie}
\end{table*}

{\color{highlight} \paragraph{Annotation Process}

The annotation guide was conducted by two medical experts from Class A tertiary hospitals and optimized during the trail annotation process. A total of 20 annotators had participated in the annotation process, including 2 medical experts who are also the owner of the annotation guideline, 2 experts from the biomedical informatics field, 4 medical M.D., and 14 master students from computer science majors. The annotation lasts for about four months (from October 2018 to December 2018), which contains the annotation time as well as the curation time. The total expense is about 40,000 RMB.

Similar to the CMeEE dataset, the annotation process for CMeIE also contains the trail annotation stage and the formal annotation stage following the same process. Besides, an additional step called the Chinese segmentation validation step was added for this dataset. The data provider has developed a segmentation tool for the medical texts which could generate the segment as well as the POS tagging, and some specified POS types (like 'disease,' 'drug') could help validate if there were potential missing named entities for this task automatically,  which could help assist the annotators to check the missing labels. The final IAA for this dataset is 0.83, which could satisfy the research purpose.
}

{\color{highlight} \paragraph{PII and IRB}

The corpus is collected from authorized medical textbooks or Clinical Practice, and no personally identifiable
information or offensive content is involved in the text. 

No PII is included in the above-mentioned resources. The dataset does not refer to ethics, which has been checked by the IRB committee of the provider.

}

\paragraph{Evaluation Metrics}
The SPO results given by the participants need to be accurately matched.
The strict Micro-F1 is used for evaluation.

\paragraph{Dataset Statistic}
This task has 14,339 training set data,  3,585 validation set data, 4,482 test set data.
The dataset is from the pediatric corpus and common disease corpus.
The pediatric corpus originates from 518 pediatric diseases, and the common disease corpus is derived from 109 common diseases.
The dataset contains nearly 75,000 triples, 28,000 disease sentences, and 53 schemas.


\paragraph{Dataset Provider}
The dataset is provided by:
\begin{itemize}
    \item Key Laboratory of Computational Linguistics, Peking University, Ministry of Education, China
    \item Laboratory of Natural Language Processing, Zhengzhou University, China
    \item The Research Center for Artificial Intelligence, Peng Cheng Laboratory, China 
    \item Harbin Institute of Technology, Shenzhen, China
\end{itemize}

\subsection{CHIP - Clinical Diagnosis Normalization Dataset (CHIP-CDN)}

\paragraph{Task Background}
Clinical term normalization is a crucial task for both research and industry use.
Clinically, there might be up to hundreds of different synonyms for the same diagnosis, symptoms, or procedures; for example, ``heart tack'' and ``MI'' both stand for the standard terminology ``myocardial infarction''. 
The goal of this task is to find the standard phrases (i.e., ICD codes) for the given clinical term.
With the help of the standard code, it can help ease the burden of researchers for the statistical analysis of clinical trials; also, it can be helpful for the insurance companies on the DRGs or DIP-related applications.
This task is proposed for this purpose, and the originally shared task was released at the CHIP2020 conference.

\paragraph{Task Description}
The task aims to standardize the terms from the final diagnoses of Chinese electronic medical records. 
No privacy information is involved in the final diagnosis.
Given the original terms, it is required to predict its corresponding standard phrase from the standard vocabulary of ``International Classification of Diseases (ICD-10) for Beijing Clinical Edition v601''. {\color{highlight}  For the detailed annotation instructions, please refer to the CBLUE official website}. Examples are shown in Table \ref{CHIP-CDN}.

\begin{table*}[!htb]
    \begin{center}
    \centering
    \begin{tabular}{l c}
    \toprule
    \textbf{Original terms} & \textbf{Normalization terms} \\
    \toprule
    \makecell[l]{
    右肺结节转移可能大\\
    Possible nodule metastasis\\
    in the right lung
    }
    & 
    \makecell[l]{
        肺占位性病变\#\# \\
        Space-occupying Lesion of the Lung\\
        肺继发恶性肿瘤\#\# \\
        Secondary Malignant Neoplasm of the Lung\\
        转移性肿瘤\\
        Metastatic Tumor
    }
    \\
    \midrule
    \makecell[l]{
    右肺结节住院\\ 
    Hospitalization after detection\\
    of nodules in the right lung
    }
    &
    \makecell[l]{
    肺占位性病变 \\
    Space-occupying Lesion of the Lung    
    }
    \\
    \midrule
    \makecell[l]{
        左上肺胸膜下结节待查 \\
        Subpleural nodule in the left\\
        upper lung to be examined
    }
    &
    \makecell[l]{
        胸膜占位 \\
        Space-occupying Lesion within the Pleural Space
    }\\
    \bottomrule
    \end{tabular}
    \end{center}
    \caption{
    Examples in CHIP-CDN
    }
    \label{CHIP-CDN}
\end{table*}

{\color{highlight} \paragraph{Annotation Process}

The Chinese Diagnostic Normalization Data Set (CHIP-CDN) was annotated by the medical team of Yidu Cloud. They are all composed of people with medical backgrounds and clinician qualification certificates. This work took about 2 months, and since the work was done by internal staff, the estimated cost was around 100,000 RMB in total.

The Chinese Diagnostic Normalization Data Set (CHIP-CDN) is completed by one round of labeling, one round of full audit, and one round of random quality inspection. Labeling and review are completed by ordinary labeling personnel with clinical qualifications, and random quality inspections are completed by high-level terminology experts.
}

{\color{highlight} \paragraph{PII and IRB}
The corpus is collected from EMR(electronic medical records), and only the final diagnoses part is chosen for research purposes. The dataset does not refer to ethics.

As shown in the example table, the final diagnosis has no PII included. 

The original dataset format is a self-defined xlsx format. To unify the data pre-processing step, the CBLUE team has converted the data format to the JSON format with the permission of the data provider.
}

\paragraph{Evaluation Metrics}
The F1 score is calculated with (original diagnosis terms, standard phrases) pairs.
Say, if the test set has $m$ golden pairs, and the predicted result has $n$ pairs, where $k$ pairs are predicted correctly, then:
\begin{equation}
    P=k/n, R=k/m, F1 = 2*P*R/(P+R).
\end{equation}

\paragraph{Dataset Statistic}
8,000 training instances and 10,000 testing instances are provided. 
We split the original training set into 6,000 and 2,000 for the training and validation set, respectively.

\paragraph{Dataset Provider}
The dataset is provided by Yidu Cloud Technology Inc.

\subsection{Clinical Trial Criterion Dataset (CHIP-CTC)}

\paragraph{Task Background}
Clinical trials refer to scientific research conducted by human volunteers to determine the efficacy, safety, and side effects of a drug or a treatment method.
It plays a crucial role in promoting the development of medicine and improving human health.
Depending on the purpose of the experiment, the subjects may be patients or healthy volunteers.
The goal of this task is to predict whether a subject meets a clinical trial or not.
Recruitment of subjects for clinical trials is generally done through manual comparison of medical records and clinical trial screening criteria, which is time-consuming, laborious, and inefficient.
In recent years, methods based on natural language processing have got successful in many biomedical applications. 
This task is proposed with the purpose of automatically classifying clinical trial eligibility criteria for the Chinese language, and the original task is released at the CHIP2019 conference.
All the data comes from real clinical trials collected from the website of the Chinese Clinical Trial Registry (ChiCTR) \footnote{\url{http://chictr.org.cn/}}, which is a non-profit organization providing registration for public research use. Each 

\paragraph{Task Description}
A total of 44 pre-defined semantic categories are defined for this task, and the goal is to predict a given text to the correct category. 
{\color{highlight}  For the detailed annotation instructions, please refer to the CBLUE official website}. Examples of labeled data are shown in Table \ref{CHIP-CTC}.
\begin{table*}[!htb]
    \begin{center}
    \centering
    \begin{tabular}{l l l}
    \toprule
    \textbf{ID} & \textbf{Clinical trial sentence} & \textbf{Category} \\
    \toprule
    S1 & 
    \makecell[l]{
        年龄>80岁\\
        Age: > 80
    }
    & Age \\
    \midrule
    S2  & 
    \makecell[l]{
        近期颅内或椎管内手术史\\
        Recent intracranial/intraspinal surgery
    }
    & Therapy or Surgery \\
    \midrule
    S3  & 
 
    \makecell[l]{
        血糖<2.7mmol/L  \\   
        Blood glucose < 2.7 mmol/L
    }
    & Laboratory Examinations 

    \\
    \bottomrule
    \end{tabular}
    \end{center}
    \caption{
    Examples in CHIP-CTC
    }
    \label{CHIP-CTC}
\end{table*}

{\color{highlight} \paragraph{Annotation Process}

The CHIP-CTC corpus was annotated by three annotators. The first annotator is Zuofeng Li, a principal scientist in Philips Research China, with more than a decade of research experience in the biomedical domain. Other annotators were Zeyu Zhang (Ph.D. candidate) and Jinxuan Yang (Ph.D. candidate) in the biomedical informatics field from Tongji University. The annotation started in July 2019 and took about 1 month. Further, the corpus was used in the CHIP 2019 shared task. The annotation was related to the annotator’s research project, and no payment was required. 

One experienced biomedical researcher (Z.L) and two raters (Z.Z and J.Y, Ph.D. candidate for biomedical informatics) of biomedical domains labeled the CHIP-CTC corpus with the 44 categories. First, they studied these categories' definitions, investigated a large number of expression patterns of criteria sentences, and chose criteria examples of each category. Next, the two raters independently annotated the same 1000 sentences, then they checked annotations and discussed contradictions with Z.L until consensus was achieved. This step repeated 20 iterations, and 20000 criteria sentences were annotated, which were later used to calculate the inter-annotator agreement score (0.9920 by Cohen’s kappa score). Finally, the remaining 18341 sentences were assigned to the two raters for annotation.
}

{\color{highlight} \paragraph{PII and IRB}
The corpus is collected from the Chinese Clinical Trial Registry (ChiCTR) website, which is a non-profit organization providing registration for public research use. For each registered clinical trial case on this website, it is already approved by the ethics committee of the organization. In addition, the annotation and corpus have also been reviewed and approved by Internal Committee on Biomedical Experiments (ICBE) in Philips. It is encouraged to use the corpus for academic research.

For each registered clinical trial report, no PII is included.

The original dataset format is a self-defined csv format. To unify the data pre-processing step, the CBLUE team has converted the data format to the JSON format with the permission of the data provider.
}

\paragraph{Evaluation Metrics}

The evaluation of this task uses Macro-F1.
Suppose we have n categories, $C_1, ..., C_i, ..., C_n$.
The accuracy rate $P_i$ is the number of records correctly predicted to class $C_i$ $/$ the number of records predicted to be class $C_i$.
Recall rate $R_i$ = the number of records correctly predicted as the class $C_i$ $/$ the number of records of the real $C_i$ class.

\begin{equation}
    Average - F1 = (1/n)\sum_{i=1}^{n} \frac{2*Pi*Ri}{Pi+Ri} 
\end{equation}

\paragraph{Dataset Statistic}

This task has 22,962 training sets, 7,682 validation sets, and 10000 test sets.

\paragraph{Dataset Provider}
The dataset is provided by the School of Life Sciences and Technology, Tongji University, and Philips Research China.

\subsection{Semantic Textual Similarity Dataset (CHIP-STS)}

\paragraph{Task Background}
CHIP-STS task aims to learn similar knowledge between disease types based on the Chinese online medical questions.
Specifically, given question pairs from 5 different diseases, it is required to determine whether the semantics of the two sentences are similar or not. 
The originally shared task was released at the CHIP2019 conference.

\paragraph{Task Description}
The category represents the name of the disease type, including diabetes, hypertension, hepatitis, aids, and breast cancer. 
The label indicates whether the semantics of the questions are the same.
If they are the same, they are marked as 1, and if they are not the same, they are marked as 0. 
Examples of labeling are shown in Table \ref{CHIP-STS}.

\begin{table*}[!htb]
    \begin{center}
    \centering
    \begin{tabular}{l l c}
    \toprule
    \textbf{Question1} & \textbf{Question2} & \textbf{Label} \\
    \toprule
    \makecell[l]{
        糖尿病吃什么？\\
        What should patients with diabetes eat?
    }
    & 
    \makecell[l]{
        糖尿病的食谱？\\
        What is the recommended dietary\\
        for patients with diabetes?
    }
    & 
    label:1 \\
    \midrule
    \makecell[l]{
        乙肝小三阳的危害？\\
        What is the harm of hepatitis B\\
        (HBsAg/HBeAb/HBcAb-positive)?
    }
    & 
    \makecell[l]{
        乙肝大三阳的危害？\\
        What is the harm of hepatitis B\\
        (HBsAg/HBeAg/HBcAb-positive)?
    }
    & label:0 \\
    \bottomrule
    \end{tabular}
    \end{center}
    \caption{
    Examples in CHIP-STS
    }
    \label{CHIP-STS}
\end{table*}

{\color{highlight} \paragraph{Annotation Process}

The CHIP-STS corpus was annotated by five undergraduate annotators from medical colleges under the guidance of one surgeon and one physician. The task is relatively simple since it is a two-class classification one; the annotation process, as well as the time of verification, lasts for two weeks. A total of 30,000 sentences pairs are annotated, and the annotation expense is 25,000 RMB.

There are five types of diseases, so each annotator was assigned two types of disease to the label to guarantee that each type of disease was annotated by two raters. During the trail annotation process, each annotator was given 100 records to label, which aimed to test if they could understand the tasks thoroughly. Following that, the annotators start to label the process, and medical experts would give necessary help, like explaining the disease mechanism to assist the raters. Finally, each record was labeled by two different labelers, and the disagreed pairs were selected for discussion and case study; the annotators would recheck the previous labeled results according to the experts' feedback. The IAA score was 0.93. 
}

{\color{highlight} \paragraph{PII and IRB}
The corpus is collected from online questions from the medical forum, and it doesn't refer to the ethics, which has been checked by the IRB committee of the provider.

During the annotation step, sentences with PHI information are discarded by the annotators manually. The CBLUE team has also validated the dataset record by record to guarantee there is no PII included.

The original dataset format is a self-defined csv format. To unify the data pre-processing step, the CBLUE team has converted the data format to the JSON format with the permission of the data provider.
}

\paragraph{Evaluation Metrics}
The evaluation of this task is Macro-F1.

\paragraph{Dataset Statistic}
This task has 16,000 training sets,  4,000 validation sets, and 10,000 tests set data.


\paragraph{Dataset Provider}
The dataset is provided by Ping An Technology.

\subsection{KUAKE-Query Intent Classification Dataset (KUAKE-QIC)}

\paragraph{Task Background}
In medical search scenarios, the understanding of query intent can significantly improve the relevance of search results.
In particular, medical knowledge is highly specialized, and classifying query intentions can also help integrate medical knowledge to enhance the performance of search results.
This task is proposed for this purpose.

\paragraph{Task Description}

There are 11 categories of medical intent labels, including diagnosis, etiology analysis, treatment plan, medical advice, test result analysis, disease description, consequence prediction, precautions, intended effects, treatment fees, and others.
{\color{highlight}  For the detailed annotation instructions, please refer to the CBLUE official website}. Examples are shown in Table \ref{KUAKE-QIC}.

\begin{table*}[!htb]
    \begin{center}
    \centering
    \begin{tabular}{l l}
    \toprule
    \textbf{Intent} & \textbf{Sentences} \\
    \toprule
    \makecell[l]{
        病情诊断\\
        disease diagnosis
    }
    & \makecell[l]{
    最近早上起来浑身无力是怎么回事？\\
    Why do I always feel weak after I get up in the morning?\\
    我家宝宝快五个月了，为什么偶尔会吐清水带？\\
    Why does my 5-month-old baby occasionally vomit clear liquid?
    }\\
    \midrule
    \makecell[l]{
        注意事项\\
        precautions
    }
    & \makecell[l]{
    哮喘应该注意些什么 \\
    What should patients with asthma pay attention to? \\
    孕妇能不能吃榴莲 \\
    Can a pregnant woman eat durians?\\
    柿子不能和什么一起吃 \\
    Which food cannot be eaten together with persimmons?\\
    糖尿病人饮食注意什么啊？\\
    What should patients with diabetes pay attention to about their diet?
    } \\
    \midrule    
    \makecell[l]{
        就医建议\\
        medical advice
    }
    & \makecell[l]{
    糖尿病该做什么检查？\\
    What examination should patients with diabetes receive?\\
    肚子疼去什么科室？\\
    Which department should patients with stomachache visit?
    } \\
    \bottomrule
    \end{tabular}
    \end{center}
    \caption{
    Examples in KUAKE-QIC 
    }
    \label{KUAKE-QIC}
\end{table*}

{\color{highlight} \paragraph{Annotation Process}

The KUAKE-QIC corpus was annotated by six annotators who graduated from medical college; they were employed by Alibaba as full-time employees for the KUAKE department. They must get past the test for the specified annotation tasks before the annotation starts. This task cost about 2 weeks, and the annotation fee was 6,600 RMB with 22,000 labeled records, that's to say, 0.3 RMB / per record.  

The annotation process was divided into three steps:

The first step was the trail annotation step; 2,000 records were selected for this stage. The annotators were grouped into 2 groups, each with 3 persons. The data provider had a strict metric for quality control, say, the IAA between the three persons within the same group must exceed 0.9. 

The second stage is the formal annotation phase, and during this stage, 6 annotators were divided into three groups, each with 2 persons. A total of 20,000 records were annotated; IAA for this step was 0.9230.

The last step was the quality control step, the sampling strategy was adopted, and 300 records were sampled for validation; some common annotation problems were raised by the medical experts, and the data would be fixed in a batch mode. In addition, some disagreed cases were made final decisions by the medical experts.

}

{\color{highlight} \paragraph{PII and IRB}
The corpus is collected from user queries from the KUAKE search engine, and it doesn't refer to the ethics, which has been checked by the IRB committee of the provider.

During the annotation step, sentences with PHI information or offensive information (like sexual queries) are discarded by the annotators manually. The dataset also got passed the data disclosure process of Alibaba.

The CBLUE team has also validated the dataset record by record to guarantee there is no PII included.
}

\paragraph{Evaluation Metrics}
Accuracy is used for the evaluation of this task.

\paragraph{Dataset Statistic}
This task has 6,931 training set data,  1,955 validation set data, and 1,994 test set data.

\paragraph{Dataset Provider}
The dataset is provided by Alibaba QUAKE Search Engine.

\subsection{KUAKE- Query Title Relevance Dataset (KUAKE-QTR)}
\paragraph{Task Background}
KUAKE Query Title Relevance is a dataset for query document (title) relevance estimation. 
For example, give the query ``Symptoms of vitamin B deficiency'', the relevant title should be ``The main manifestations of vitamin B deficiency''.

\paragraph{Task Description}
The correlation between Query and Title is divided into 4 levels (0-3), 0 is the worst, and 3 stands for the best match.
{\color{highlight}  For the detailed annotation instructions, please refer to the CBLUE official website}. Examples are shown in Table \ref{KUAKE-QTR}.

\begin{table*}[!htb]
    \begin{center}
    \centering
    \begin{tabular}{l l l}
    \toprule
    \textbf{Query} & \textbf{Title} & \textbf{Level} \\
    \toprule
    \makecell[l]{
        缺维生素b的症状\\
        Symptoms of Vitamin B deficiency
    }
    & 
    \makecell[l]{
        维生素b缺乏症的主要表现\\ 
        What are the major symptoms of \\
        Vitamin B deficiency?
    }

    & 3 \\
    \midrule
    \makecell[l]{
        大腿软组织损伤怎么办\\
        How can I treat a soft tissue\\
        injury in the thigh?
    }
    & 
    \makecell[l]{
        腿部软组织损伤怎么办\\
        What's the treatment for a soft tissue\\
        injury in the leg?
    }
    & 2 \\
    \midrule
    \makecell[l]{
        小腿抽筋是什么原因引起的\\
        What causes lower leg cramps?
    }
    & 
    \makecell[l]{
        小腿抽筋后一直疼怎么办\\
        How can I treat pains caused by lower\\
        leg cramps?
    }    
    & 1 \\
    \makecell[l]{
        挑食是什么原因造成的\\
        What is the cause of picky eating?
    }&
    \makecell[l]{
        挑食是什么原因造成的\\
        What is the cause of picky eating?
    }    
    &
    0 \\
    \bottomrule
    \end{tabular}
    \end{center}
    \caption{
    Examples in KUAKE-QTR
    }
    \label{KUAKE-QTR}
\end{table*}

{\color{highlight} \paragraph{Annotation Process}

The KUAKE-QTR corpus was annotated by a total of nine annotators, among which seven were from third-party crowd-sourcing undergraduates from medical colleges, and two were from Alibaba full-time employees with medical backgrounds. The crowd-sourcing annotators were required to get trained and pass the annotation test before they could execute the task. The annotations lasted for 2 weeks, and a total of 28,000 RMB was used.

Similar to the KUAKE-QIC task, the KUAKE-QTR annotation process was divided into three steps with minor changes:

The training and examination stage: The seven annotators got trained by the two FTE (full-time employee) experts to understand the tasks, then each one was given 200 records to label, which have ground-truth answer annotated by FTE experts. The precision must be above 85\%  to get past the test.

The second step was the formal annotation step, and Each annotator was given 3,000 records to label, among which 100 were with golden labels. The annotation tools would automatically evaluate the annotation quality by comparing the label between the annotators' ones and the golden ones. Help would be given to the annotators if necessary. Only the precision exceeding the threshold 0.85 would be handed to the next round.

The last step was the quality control step, the sampling strategy was adopted, and 100 records were sampled for validation by the FTE medical experts; bad cases would be returned to the crowd-sourcing annotators to be fixed.
}

{\color{highlight} \paragraph{PII and IRB}
The corpus is collected from user queries from the KUAKE search engine, and it doesn't refer to the ethics, which has been checked by the IRB committee of the provider.

During the annotation step, sentences with PHI information or offensive information (like sexual queries) are discarded by the annotators manually. The dataset also got passed the data disclosure process of Alibaba.

The CBLUE team has also validated the dataset record by record to guarantee there is no PII included. One record with the NULL label was discarded with the permission of the provider.
}

\paragraph{Evaluation Metrics}
Same as the KUAKE-QIC task, accuracy is used for the evaluation of this task.

\paragraph{Dataset Statistic}

This task has 24,174 training set data, 2,913 validation set data, and 54,65 test set data.


\paragraph{Dataset Provider}
This dataset is provided by Alibaba QUAKE Search Engine.

\subsection{KUAKE - Query Query Relevance Dataset (KUAKE-QQR)}
\paragraph{Task Background}
KUAKE Query-Query Relevance is a dataset that evaluates the relevance between two given queries to resolve the long-tail challenges for search engines.
Similar to KUAKE-QTR, query-query relevance is an essential and challenging task in real-world search engines.

\paragraph{Task Description}
The correlation between Query and Title is divided into 3 levels (0-2), 0 is the worst, and 2 stands for the best correlation. {\color{highlight}  For the detailed annotation instructions, please refer to the CBLUE official website}. Examples are shown in Table \ref{KUAKE-QQR-example}.

\begin{table*}[!htb]
    \begin{center}
    \centering
    \begin{tabular}{l l l}
    \toprule
    \textbf{Query} & \textbf{Query} & \textbf{Level} \\
    \toprule
    \makecell[l]{
        小孩子打呼噜是什么原因引起的\\
        What causes children's snoring
    }
    & 
    \makecell[l]{
        小孩子打呼噜什么原因\\
        What makes children snore?
    }    
    & 2 \\
    \midrule
    \makecell[l]{
        双眼皮遗传规律\\
        Heredity laws of double-fold eyelids
    }
    & 
    \makecell[l]{
       内双眼皮遗传\\
       Heredity of hidden double-fold eyelids
    }
    & 1 \\
    \midrule
    \makecell[l]{
        白血病血常规有啥异常\\
        What index of the CBC test will be abnormal for \\
        patients with leukemia?
    }
    & 
    \makecell[l]{
        白血病血检有哪些异常\\
        What index of the blood test will be abnormal\\
        for patients with leukemia?
    }
    & 0 \\
    \bottomrule
    \end{tabular}
    \end{center}
    \caption{
    Examples in KUAKE-QQR
    }
    \label{KUAKE-QQR-example}
\end{table*}

{\color{highlight} \paragraph{Annotation Process}
The same as KUAKE-QTR except for the expense, which is 22,000 RMB in total.

}

{\color{highlight} \paragraph{PII and IRB}
The same as KUAKE-QTR.
}

\paragraph{Evaluation Metrics}
Same with the KUAKE-QIC and KUAKE-QTR tasks, accuracy is used for the evaluation metrics.

\paragraph{Dataset Statistic}

This task has 15,000 training set data, 1,600 validation set data,  and 1,596 test set data.


\paragraph{Dataset Provider}
This dataset is provided by Alibaba QUAKE Search Engine.

\section{Experiments Details}
This section details the training procedures and hyper-parameters for each of the data sets. 
We utilize Pytorch to conduct experiments, and all running hyper-parameters are shown in the following Tables. 
There are two stages in CMeIE, namely, entity recognition (CMeEE-ER) and relation classification (CMeEE-RE). So we detail the hyper-parameters in  CMeEE-ER and CMeEE-RE, respectively.

\textbf{Requirements}
    \begin{itemize}
        \item python3
        \item pytorch 1.7
        \item transformers 4.5.1
        \item jieba
        \item gensim
    \end{itemize}

\textbf{Hyper-parameters for Specific Task} is shown in Table \ref{common}-\ref{hyper-KUAKE-QQR}

\begin{table*}[]
    \begin{center}
    \centering
    \begin{tabular}{c c}
    \toprule
    \tf{Method} & \tf{Value} \\
    \toprule
    
    warmup\_proportion &	0.1 \\
    weight\_decay & 0.01 \\
    adam\_epsilon & 1e-8 \\
    max\_grad\_norm & 1.0 \\
        
    \bottomrule
    \end{tabular}
    \end{center}
    \caption{
    Common hyper-parameters for all CBLUE tasks
    }
    \label{common}
\end{table*}

\begin{table*}[]
    \begin{center}
    \centering
    \begin{tabular}{l c c c c}
    \toprule
    \tf{Model} & \tf{epoch} & \tf{batch\_size} & \tf{max\_length} & \tf{learning\_rate} \\
    \toprule
    bert-base & 5 & 32 & 128 & 4e-5 \\
    bert-wwm-ext & 5 & 32 & 128 & 4e-5 \\
    roberta-wwm-ext & 5 & 32 & 128 & 4e-5 \\
    roberta-wwm-ext-large & 5 & 12 & 65 & 2e-5 \\
    roberta-large & 5 & 12 & 65 & 2e-5 \\
    albert-tiny & 10 & 32 & 128 & 5e-5 \\
    albert-xxlarge & 5 & 12 & 65 & 1e-5 \\
    zen & 5 & 20 & 128 & 4e-5 \\
    macbert-base & 5 & 32 & 128 & 4e-5 \\
    macbert-large & 5 & 12 & 80 & 2e-5 \\
    PCL-MedBERT & 5 & 32 & 128 & 4e-5 \\
    \bottomrule
    \end{tabular}
    \end{center}
    \caption{
    Hyper-parameters for the training of pre-trained models with a token classification head on top for named entity recognition of the CMeEE task.
    }
\end{table*}

\begin{table*}[]
    \begin{center}
    \centering
    \begin{tabular}{l c c c c}
    \toprule
    \tf{Model} & \tf{epoch} & \tf{batch\_size} & \tf{max\_length} & \tf{learning\_rate} \\
    \toprule
    bert-base & 7 & 32 & 128 & 5e-5 \\
    bert-wwm-ext & 7 & 32 & 128 & 5e-5 \\
    roberta-wwm-ext & 7 & 32 & 128 & 4e-5 \\
    roberta-wwm-ext-large & 7 & 16 & 80 & 4e-5 \\
    roberta-large & 7 & 16 & 80 & 2e-5 \\
    albert-tiny & 10 & 32 & 128 & 4e-5 \\
    albert-xxlarge & 7 & 16 & 80 & 1e-5 \\
    zen & 7 & 20 & 128 & 4e-5 \\
    macbert-base & 7 & 32 & 128 & 4e-5 \\
    macbert-large & 7 & 20 & 80 & 2e-5 \\
    PCL-MedBERT & 7 & 32 & 128 & 4e-5 \\
    \bottomrule
    \end{tabular}
    \end{center}
    \caption{
    Hyper-parameters for the training of pre-trained models with a token-level classifier for subject and object recognition of the CMeIE task.
    }
\end{table*}

\begin{table*}[]
    \begin{center}
    \centering
    \begin{tabular}{l c c c c}
    \toprule
    \tf{Model} & \tf{epoch} & \tf{batch\_size} & \tf{max\_length} & \tf{learning\_rate} \\
    \toprule
    bert-base & 8 & 32 & 128 & 5e-5 \\
    bert-wwm-ext & 8 & 32 & 128 & 5e-5 \\ 
    roberta-wwm-ext & 8 & 32 & 128 & 4e-5 \\
    roberta-wwm-ext-large & 8 & 16 & 80 & 4e-5\\
    roberta-large & 8 & 16 & 80 & 2e-5\\
    albert-tiny & 10 & 32 & 128 & 4e-5\\
    albert-xxlarge & 8 & 16 & 80 & 1e-5\\ 
    zen & 8 & 20 & 128 & 4e-5 \\
    macbert-base & 8 & 32 & 128 & 4e-5 \\
    macbert-large & 8 & 20 & 80 & 2e-5 \\
    PCL-MedBERT & 8 & 32 & 128 & 4e-5\\
    \bottomrule
    \end{tabular}
    \end{center}
    \caption{
    Hyper-parameters for the training of pre-trained models with a classifier for the entity pairs relation prediction of the CMeIE task.
    }
\end{table*}

\begin{table*}[]
    \begin{center}
    \centering
    \begin{tabular}{l c c c c}
    \toprule
    \tf{Model} & \tf{epoch} & \tf{batch\_size} & \tf{max\_length} & \tf{learning\_rate} \\
    \toprule
    bert-base & 5 & 32 & 128 & 5e-5  \\ 
    bert-wwm-ext & 5 & 32 & 128 & 5e-5  \\ 
    roberta-wwm-ext & 5 & 32 & 128 & 4e-5  \\ 
    roberta-wwm-ext-large & 5 & 20 & 50 & 3e-5  \\ 
    roberta-large & 5 & 20 & 50 & 4e-5  \\ 
    albert-tiny & 10 & 32 & 128 & 4e-5  \\ 
    albert-xxlarge & 5 & 20 & 50 & 1e-5  \\ 
    zen & 5 & 20 & 128 & 4e-5 \\
    macbert-base & 5 & 32 & 128 & 4e-5 \\
    macbert-large & 5 & 20 & 50 & 2e-5 \\
    PCL-MedBERT & 5 & 32 & 128 & 4e-5  \\ 
    \bottomrule
    \end{tabular}
    \end{center}
    \caption{
    Hyper-parameters for the training of pre-trained models with a sequence classification head on top for screening criteria classification of the CHIP-CTC task.
    }
\end{table*}

\begin{table*}[]
    \begin{center}
    \centering
    \begin{tabular}{c c}
    \toprule
    \tf{Param} & \tf{Value} \\
    \toprule
    
    recall\_k & 200 \\
    num\_negative\_sample & 10 \\
        
    \bottomrule
    \end{tabular}
    \end{center}
    \caption{
    Hyper-parameters for the CHIP-CDN task. We model the CHIP-CDN task with two stages: recall stage and ranking stage. \textit{num\_negative\_sample} sets the number of negative samples sampled for the training ranking model during the ranking stage. \textit{recall\_k} sets the number of candidates recalled in the recall stage.
    }
\end{table*}

\begin{table*}[]
    \begin{center}
    \centering
    \begin{tabular}{l c c c c}
    \toprule
    \tf{Model} & \tf{epoch} & \tf{batch\_size} & \tf{max\_length} & \tf{learning\_rate} \\
    \toprule
    bert-base & 3 & 32 & 128 & 4e-5\\ 
    bert-wwm-ext & 3 & 32 & 128 & 5e-5\\ 
    roberta-wwm-ext & 3 & 32 & 128 & 4e-5\\ 
    roberta-wwm-ext-large & 3 & 32 & 40 & 4e-5\\ 
    roberta-large & 3 & 32 & 40 & 4e-5\\ 
    albert-tiny & 3 & 32 & 128 & 4e-5\\ 
    albert-xxlarge & 3 & 32 & 40 & 1e-5\\ 
    zen & 3 & 20 & 128 & 4e-5 \\
    macbert-base & 3 & 32 & 128 & 4e-5 \\
    macbert-large & 3 & 32 & 40 & 2e-5 \\
    PCL-MedBERT & 3 & 32 & 128 & 4e-5\\
    \bottomrule
    \end{tabular}
    \end{center}
    \caption{
    Hyper-parameters for the training of pre-trained models with a sequence classifier for the ranking model of the CHIP-CDN task. We encode the pairs of the original term and standard phrase from candidates recalled during the recall stage and then pass the pooled output to the classifier, which predicts the relevance between the original term and standard phrase.
    }
\end{table*}

\begin{table*}[]
    \begin{center}
    \centering
    \begin{tabular}{l c c c c}
    \toprule
    \tf{Model} & \tf{epoch} & \tf{batch\_size} & \tf{max\_length} & \tf{learning\_rate} \\
    \toprule
    bert-base & 20 & 32 & 128 & 4e-5 \\ 
    bert-wwm-ext & 20 & 32 & 128 & 5e-5 \\ 
    roberta-wwm-ext & 20 & 32 & 128 & 4e-5 \\ 
    roberta-wwm-ext-large & 20 & 12 & 40 & 4e-5 \\ 
    roberta-large & 20 & 12 & 40 & 4e-5 \\ 
    albert-tiny & 20 & 32 & 128 & 4e-5 \\ 
    albert-xxlarge & 20 & 12 & 40 & 1e-5 \\ 
    zen & 20 & 20 & 128 & 4e-5 \\
    macbert-base & 20 & 32 & 128 & 4e-5 \\
    macbert-large & 20 & 12 & 40 & 2e-5 \\
    PCL-MedBERT & 20 & 32 & 128 & 4e-5 \\ 
    \bottomrule
    \end{tabular}
    \end{center}
    \caption{
    Hyper-parameters for the training of pre-trained models with a sequence classifier for the prediction of the number of standard phrases corresponding to the original term in the CHIP-CDN task. 
    }
\end{table*}

\begin{table*}[]
    \begin{center}
    \centering
    \begin{tabular}{l c c c c}
    \toprule
    \tf{Model} & \tf{epoch} & \tf{batch\_size} & \tf{max\_length} & \tf{learning\_rate} \\
    \toprule
    bert-base & 3 & 16 & 40 & 3e-5\\ 
    bert-wwm-ext & 3 & 16 & 40 & 3e-5\\ 
    roberta-wwm-ext & 3 & 16 & 40 & 4e-5\\ 
    roberta-wwm-ext-large & 3 & 16 & 40 & 4e-5\\ 
    roberta-large & 3 & 16 & 40 & 2e-5\\ 
    albert-tiny & 3 & 16 & 40 & 5e-5\\ 
    albert-xxlarge & 3 & 16 & 40 & 1e-5\\
    zen & 3 & 16 & 40 & 2e-5 \\
    macbert-base & 3 & 16 & 40 & 3e-5 \\
    macbert-large & 3 & 16 & 40 & 3e-5 \\
    PCL-MedBERT & 3 & 16 & 40 & 2e-5\\ 
    \bottomrule
    \end{tabular}
    \end{center}
    \caption{
    Hyper-parameters for the training of pre-trained models with a sequence classifier for sentence similarity predication of the CHIP-STS task.
    }
\end{table*}

\begin{table*}[]
    \begin{center}
    \centering
    \begin{tabular}{l c c c c}
    \toprule
    \tf{Model} & \tf{epoch} & \tf{batch\_size} & \tf{max\_length} & \tf{learning\_rate} \\
    \toprule
    bert-base & 3 & 16 & 50 & 2e-5 \\ 
    bert-wwm-ext & 3 & 16 & 50 & 2e-5 \\ 
    roberta-wwm-ext & 3 & 16 & 50 & 2e-5 \\ 
    roberta-wwm-ext-large & 3 & 16 & 50 & 2e-5 \\ 
    roberta-large & 3 & 16 & 50 & 3e-5 \\ 
    albert-tiny & 3 & 16 & 50 & 5e-5 \\ 
    albert-xxlarge & 3 & 16 & 50 & 1e-5 \\ 
    zen & 3 & 16 & 50 & 2e-5 \\
    macbert-base & 3 & 16 & 50 & 3e-5 \\
    macbert-large & 3 & 16 & 50 & 2e-5 \\
    PCL-MedBERT & 3 & 16 & 50 & 2e-5 \\ 
    \bottomrule
    \end{tabular}
    \end{center}
    \caption{
    Hyper-parameters for the training of pre-trained models with a sequence classifier for query intention prediction of the KUAKE-QIC task.
    }
\end{table*}

\begin{table*}[]
    \begin{center}
    \centering
    \begin{tabular}{l c c c c}
    \toprule
    \tf{Model} & \tf{epoch} & \tf{batch\_size} & \tf{max\_length} & \tf{learning\_rate} \\
    \toprule
    
    bert-base & 3 & 16 & 40 & 4e-5\\ 
    bert-wwm-ext & 3 & 16 & 40 & 2e-5\\ 
    roberta-wwm-ext & 3 & 16 & 40 & 3e-5\\ 
    roberta-wwm-ext-large & 3 & 16 & 40 & 2e-5\\ 
    roberta-large & 3 & 16 & 40 & 2e-5\\ 
    albert-tiny & 3 & 16 & 40 & 5e-5\\ 
    albert-xxlarge & 3 & 16 & 40 & 1e-5\\ 
    zen & 3 & 16 & 40 & 3e-5 \\
    macbert-base & 3 & 16 & 40 & 2e-5 \\
    macbert-large & 3 & 16 & 40 & 2e-5 \\
    PCL-MedBERT & 3 & 16 & 40 & 3e-5\\ 

    \bottomrule
    \end{tabular}
    \end{center}
    \caption{
    Hyper-parameters of training the sequence classifier for the KUAKE-QTR task.
    }
    \label{hyper-KUAKE-QTR}
\end{table*}

\begin{table*}[]
    \begin{center}
    \centering
    \begin{tabular}{l c c c c}
    \toprule
    \tf{Model} & \tf{epoch} & \tf{batch\_size} & \tf{max\_length} & \tf{learning\_rate} \\
    \toprule

    bert-base & 3 & 16 & 30 & 3e-5\\ 
    bert-wwm-ext & 3 & 16 & 30 & 3e-5\\ 
    roberta-wwm-ext & 3 & 16 & 30 & 3e-5\\ 
    roberta-wwm-ext-large & 3 & 16 & 30 & 3e-5\\ 
    roberta-large & 3 & 16 & 30 & 2e-5\\ 
    albert-tiny & 3 & 16 & 30 & 5e-5\\ 
    albert-xxlarge & 3 & 16 & 30 & 3e-5\\ 
    zen & 3 & 16 & 30 & 2e-5 \\
    macbert-base & 3 & 16 & 30 & 2e-5 \\
    macbert-large & 3 & 16 & 30 & 2e-5 \\
    PCL-MedBERT & 3 & 16 & 30 & 2e-5\\ 
    
    \bottomrule
    \end{tabular}
    \end{center}
    \caption{
    Hyper-parameters of training the sequence classifier for the KUAKE-QQR task.
    }
    \label{hyper-KUAKE-QQR}
\end{table*}

\section{Error Analysis for Other Tasks}

We introduce the error definition as follows and illustrate some error cases for other tasks in Table   \ref{err1} to  \ref{err6}.

\textbf{Ambiguity} indicates that the instance has a similar context but different meaning, which mislead the prediction.

\textbf{Need domain knowledge} indicates that there exist biomedical terminologies in the instance which require domain knowledge to understand. 

\textbf{Need syntactic knowledge} indicates that there exists complex syntactic structure in the instance, and the model fails to understand the correct meaning. 

\textbf{Entity overlap } indicates there exist multiple overlapping entities in the instance. 

\textbf{Long sequence} indicates that the input instance is very long. 

\textbf{Annotation error} indicates that the annotated label is wrong. 

\textbf{Wrong entity boundary} indicates that the instance has the wrong entity boundary.

\textbf{Rare words} indicates that there exist low-frequency words in the instance.

\textbf{Multiple triggers} indicates that there exist multiple indicative words which mislead the prediction.  

\textbf{Colloquialism} (very common in the search queries) indicates that the instance is quite different from written language (e.g., with many abbreviations), thus, challenging the prediction model.

\textbf{Irrelevant description} indicates that the instance has lots of irrelevant information, which mislead the prediction.

\begin{CJK*}{UTF8}{gbsn}
\begin{table*}[]
  \footnotesize
  \centering
  \setlength\tabcolsep{4pt}
  \label{tab:cmeie_badcase}
  \begin{tabular}{p{3.5cm}p{3cm}<{\centering}p{3cm}<{\centering}p{3cm}<{\centering}}
    \toprule
    {\textbf{Sentence}} &
    {\textbf{Golden}} &
    {\textbf{RO}} &
    {\textbf{ME}} \\
    \midrule
    另一项研究显示，减荷鞋对内侧膝骨关节炎也没有效。&
    内侧膝骨关节炎 | 辅助治疗 | 减荷鞋 & 膝骨关节炎|辅助治疗 |减荷鞋 & 膝骨关节炎| 辅助治疗|减荷鞋 \\
    Another study showed that load-reducing shoes were not effective for medial knee osteoarthritis. & 
    medial knee osteoarthritis, adjuvant therapy, load-reducing shoes
    & medial knee osteoarthritis, adjuvant therapy, load-reducing shoes
    & medial knee osteoarthritis, adjuvant therapy, load-reducing shoes \\
    \midrule    
    精神疾病：焦虑和抑郁与失眠症高度相关。& 
    焦虑|相关（导致）|失眠症 & 无|无|无 & 焦虑|相关（导致）|失眠症 \\
    Mental illness: anxiety and depression are related to insomnia. & 
    anxiety, related cause, insomnia & None|None|None & anxiety, related cause, insomnia\\
    \midrule
    在狂犬病感染晚期，患者常出现昏迷。& 
    狂犬病|相关（转化）|昏迷 & 无|无|无 & 无|无|无 \\
    In the late stage of rabies infection, patients often appear comatose. & rabies, transform, comatose & None|None|None & None|None|None \\
    \bottomrule
  \end{tabular}
  \caption{
  Error cases in CMeIE.
  We evaluate roberta-wwm-ext and PCL-MedBERT on 3 sampled sentences, with their gold labels and model predictions.
  Each label consists of subject | predicate | Object. 
  None means that the model fails to predict. RO = roberta-wwm-ext, MB = PCL-MedBERT.}
  \label{err1}
\end{table*}
\end{CJK*}

\begin{CJK*}{UTF8}{gbsn}
\begin{table*}[]
  \footnotesize
  \centering
  \setlength\tabcolsep{4pt}
  \label{tab:cdn_badcase}
  \begin{tabular}{p{4.5cm}p{2.8cm}<{\centering}p{2.8cm}<{\centering}p{2.8cm}<{\centering}}
    \toprule
    {\textbf{Sentence}} &
    {\textbf{Label}} & 
    {\textbf{RO}} & 
    {\textbf{MB}} \\
    \midrule
    右第一足趾创伤性足趾切断 & 单趾切断 & 足趾损伤 & 单趾切断\\
    Right first toe traumatic toe cutting & Single toe cut & Toe injury & Single toe cut\\
    \midrule
    C3-4脊髓损伤 & 颈部脊髓损伤 & 脊髓损伤 & 脊髓损伤\\
    C3-4 spinal cord injury & Neck spinal cord injury & Spinal cord injury & Spinal cord injury \\
    \midrule
    肿瘤骨转移胃炎 & 骨继发恶性肿瘤\#\#转移性肿瘤\#\#胃炎 & 反流性胃炎\#\#转移性肿瘤\#\#胃炎 & 骨盆部肿瘤\#\#转移性肿瘤\#\#胃炎\\
    Tumor bone metastatic gastritis & Junior malignant tumor\#\#Metastatic tumor\#\#Gastritis & Reflux gastritis\#\#Metastatic tumor\#\#Gastritis & Pelvic tumor\#\#Metastatic tumor\#\#Gastritis \\
    \bottomrule
  \end{tabular}
  \caption{
  Error cases in CHIP-CDN. 
  We evaluate roberta-wwm-ext and PCL-MedBERT on 3 sampled sentences, with their gold labels and model predictions. There may be multiple predicted values, separated by a "\#\#". RO = roberta-wwm-ext, MB = PCL-MedBERT.}
    \label{err2}
\end{table*}
\end{CJK*}

\begin{CJK*}{UTF8}{gbsn}
\begin{table*}[]
  \footnotesize
  \centering
  \setlength\tabcolsep{4pt}
  \label{tab:ctc_badcase}
   \centering
  \begin{tabular}{p{5.5cm}ccc}
    \toprule
    {\textbf{Sentence}} &
    {\textbf{Label}} & 
    {\textbf{RO}} & 
    {\textbf{MB}} \\
    \midrule
    既往多次行剖腹手术或腹腔广泛粘连者 & 含有多类别的语句 & 治疗或手术 & 治疗或手术\\
    Previous multi-time crashed surgery or abdominal adhesive & Multiple & Therapy or Surgery & Therapy or Surgery\\
    \midrule
    术前认知发育筛查（DST）发现发育迟缓 & 诊断 & 疾病 & 诊断\\
    Preoperative cognitive development screening test(DST) finds development slow & Diagnostic & Disease & Diagnostic \\
    \midrule
    已知发生中枢神经系统转移的患者 & 肿瘤进展 & 疾病 & 疾病\\
    Patients who have been transferred in central nervous system & Neoplasm Status & Disease & Disease \\
    \bottomrule
  \end{tabular}
  \caption{
  Error cases in CHIP-CTC. 
  We evaluate roberta-wwm-ext and PCL-MedBERT on 3 sampled sentences, with their gold labels and model predictions. RO = roberta-wwm-ext, MB = PCL-MedBERT.}
    \label{err3}
\end{table*}
\end{CJK*}

\begin{CJK*}{UTF8}{gbsn}
\begin{table*}[]
  \footnotesize
  \label{tab:sts_badcase}
  \centering
  \begin{tabular}{p{4.2cm}p{4.2cm}cccc}
    \toprule
    \multirow{2}{*}{\textbf{Query-A}} &
    \multirow{2}{*}{\textbf{Query-B}} & 
    \multicolumn{3}{c}{\textbf{Model}} &
    \multirow{2}{*}{\textbf{Gold}} \\
    & & BE & BE+ & MB & \\
    \midrule
    汗液能传播乙肝病毒吗？ & 乙肝的传播途径？ & 0 & 0 & 0 & 1 \\
    Can sweat spread the hepatitis B virus? & 
    How is hepatitis B transmitted? & & & & \\
    \midrule
    哪种类型糖尿病？& 我是什么类型的糖尿病？ & 1 & 1 & 1 & 0 \\
    What type of diabetes? & What type of diabetes am I? & & & & \\
    \midrule
    如何防治艾滋病？ & 艾滋病防治条例。& 1 & 0 & 0 & 1 \\
    How to prevent AIDS? & AIDS Prevention and Control Regulations.  & & & & \\
    \bottomrule
  \end{tabular}
  \caption{Error cases in CHIP-STS. 
  We evaluate performance of baselines with 3 sampled instances.
  The similarity between queries is divided into 2 levels (0-1), which means '\textit{unrelated}' and '\textit{related}'. BE = BERT-base, BE+ = BERT-wwm-ext-base, MB = PCL-MedBERT.}
    \label{err4}
\end{table*}
\end{CJK*}

\begin{CJK*}{UTF8}{gbsn}
\begin{table*}[]
  \footnotesize
  \label{tab:qtr_badcase}
  \centering
  \begin{tabular}{p{4.2cm}p{4.2cm}cccc}
    \toprule
    \multirow{2}{*}{\textbf{Query-A}} &
    \multirow{2}{*}{\textbf{Query-B}} & 
    \multicolumn{3}{c}{\textbf{Model}} &
    \multirow{2}{*}{\textbf{Gold}} \\
    & & BE & BE+ & MB & \\
    \midrule
    吃药能吃螃蟹吗？ & 你好，吃完螃蟹后，可不可以吃药呢 & 3 & 3 & 3 & 0 \\
    Can I eat crabs with medicine? & 
    Hello, does it matter to take medicine after eating crabs? & & & & \\
    \midrule
    一颗蛋白卡路里。 & 一个鸡蛋白的热量。 & 1 & 1 & 0 & 3 \\
    Calories per egg white. & One egg white calories. & & & & \\
    \midrule
    氨基酸用法用量。 & 氨基酸的功效及用法用量。 & 2 & 2 & 2 & 1 \\
    Amino acid usage and dosage. & Efficacy and dosage of amino acids.  & & & & \\
    \bottomrule
  \end{tabular}
  \caption{Error cases in KUAKE-QTR. 
  We evaluate performance of baselines with 3 sampled instances.
  The correlation between Query and Title is divided into 4 levels (0-3), which means '\textit{unrelated}', '\textit{poorly related}', '\textit{related}' and '\textit{strongly related}'. BE = BERT-base, BE+ = BERT-wwm-ext-base, MB = PCL-MedBERT.}
    \label{err5}
\end{table*}
\end{CJK*}

\begin{CJK*}{UTF8}{gbsn}
\begin{table*}[]
  \footnotesize
  \centering
  \begin{tabular}{p{4.2cm}p{4.2cm}cccc}
    \toprule
    \multirow{2}{*}{\textbf{Query-A}} &
    \multirow{2}{*}{\textbf{Query-B}} & 
    \multicolumn{3}{c}{\textbf{Model}} &
    \multirow{2}{*}{\textbf{Gold}} \\
    & & BE & ZEN & MB & \\
    \midrule
    益生菌是饭前喝还是饭后喝。\newline Should probiotics be drunk before or after meals. & 益生菌是饭前喝还是饭后喝比较好。\newline Is it better to drink probiotics before or after meals & 1 & 2 & 1 & 2 \\
    \midrule
    糖尿病能吃肉吗？\newline Can diabetics eat meat? & 高血糖能吃肉吗?\newline Can hyperglycemic patients eat meat? & 1 & 1 & 1 & 0 \\
    \midrule
    神经衰弱吃什么药去根？ \newline What drug does neurasthenic patient take effective? & 神经衰弱吃什么药有效？\newline What drug does neurasthenic patient take effective? & 0 & 0 & 2 & 2 \\
    \bottomrule`
  \end{tabular}
  \caption{Error cases in KUAKE-QQR. We evaluate performance of baselines with 3 sampled instances.
  The correlation between Query and Title is divided into 3 levels (0-2), which means '\textit{poorly related or unrelated}', '\textit{related}' and '\textit{strongly related}'. BE = BERT-base, ZEN = ZEN, MB = PCL-MedBERT.}
   \label{err6}
\end{table*}
\end{CJK*}

\end{CJK*}

\newpage
\section*{Contributions}

\textbf{Ningyu Zhang, Zhen Bi, Xiaozhuan Liang, Lei Li} from Zhejiang University, AZFT Joint Lab for
Knowledge Engine, Hangzhou Innovation Center wrote the paper.

\textbf{Mosha Chen, Chuanqi Tan, Fei Huang, Luo Si} from Alibaba Group and \textbf{Zheng Yuan} from the Center for Statistical Science, Tsinghua University contributed the CBLUE benchmark leaderboard and transformed the eight datasets from self-defined data format to unified JSON format.

\textbf{Kunli Zhang} from School of Information Engineering, Zhengzhou University, Peng Cheng Laboratory, China and \textbf{Baobao Chang} from Key Laboratory of Computational Linguistics, Ministry of Education, Peking University, Peng Cheng Laboratory, China contributed the dataset of CMeEE.

\textbf{Hongying Zan} from School of Information Engineering, Zhengzhou University, Peng Cheng Laboratory, China and \textbf{Zhifang Sui} from Key Laboratory of Computational Linguistics, Ministry of Education, Peking University, Peng Cheng Laboratory, China contributed the dataset of CMeIE.

\textbf{Linfeng Li, Jun Yan} from Yidu Cloud Technology Inc., Beijing, China contributed the dataset of CHIP-CDN.

\textbf{Hui Zong} from School of Life Sciences and Technology, Tongji University and Philips Research China contributed the dataset of CHIP-CTC.

\textbf{Yuan Ni} from Pingan Health Technology, Shanghai, China and \textbf{Guotong Xie} from Pingan Health Technology, China, Ping An Health Cloud Company Limited, China, Ping An International Smart City Technology Co., Ltd, China contributed the dataset of CHIP-STS.

\textbf{Kangping Yin, Jian Xu} from Alibaba Group and \textbf{Xin Shang} from School of Mathematical Science, Zhejiang University contributed the datasets of KUAKE-QIC, KUAKE-QTR, and KUAKE-QQR.

\textbf{Buzhou Tang}, \textbf{Qingcai Chen} from Harbin Institute of Technology (Shenzhen), Peng Cheng Laboratory, China advised the project, suggested tasks, and led the research.

\end{document}

%% file: header.tex
\newcommand{\tf}[1]{\textbf{#1}}